\documentclass[journal]{IEEEtran}

\usepackage{graphicx}
\ifCLASSOPTIONcompsoc
\usepackage[caption=false,font=normalsize,labelfon
t=sf,textfont=sf]{subfig}
\else
\usepackage[caption=false,font=footnotesize]{subfig}
\fi
\usepackage{times} 
\usepackage{amsmath} 
\usepackage{eqnarray} 
\usepackage{amssymb}  
\usepackage{eucal}

\graphicspath{{figures/}}

\title{\LARGE \bf
Communication Model-Task Pairing in Artificial Swarm Design
}

\author{Musad~Haque$^1$, ~Connor~McGowan$^2$, ~Yifan~Guo$^2$, ~Douglas~Kirkpatrick$^3$,~and~Julie~A.~Adams$^{2,4}$
\thanks{1 Johns Hopkins University Applied Physics Laboratory}
\thanks{Laurel, MD, 20723, USA}
\thanks{\texttt{musad.haque@jhuapl.edu}}
\thanks{2 Electrical Engineering and Computer Science, Vanderbilt University}
\thanks{3 Computer Science and Engineering, Michigan State University}
\thanks{4 Electrical Engineering and Computer Science, Oregon State University}
}

\begin{document}

\maketitle

\begin{abstract}
Unraveling the nature of the communication model that governs which two individuals in a swarm interact with each other is an important line of inquiry in the collective behavior sciences. A number of models have been proposed in the biological swarm literature, with the leading models being the metric, topological, and visual models. The hypothesis evaluated in this manuscript is whether the choice of a communication model impacts the performance of a tasked artificial swarm. The biological models are used to design coordination algorithms for a simulated swarm, which are evaluated over a range of six swarm robotics tasks. Each task has an associated set of performance metrics that are used to evaluate how the communication models fare against each other. The general findings demonstrate that the communication model significantly affects the swarm's performance for individual tasks, and this result implies that the communication model-task pairing is an important consideration when designing artificial swarms. Further analysis of each tasks' performance metrics reveal instances in which pairwise considerations of model and one of the various experimental factors becomes relevant. The reported research demonstrates that the artificial swarm's task performance can be increased through the careful selection of a communications model.
\end{abstract}


\IEEEpeerreviewmaketitle


\section{Introduction}
\label{sec:introduction}


Numerous advantages are shared by animals that live in groups \cite{krauseRuxton}, which includes the ``many-eyes effect" against predators and utilizing group hunting techniques during foraging. These benefits are attributed to the coordination amongst group members. A high degree of coordination is displayed by some social animals during cooperative food retrieval \cite{prattTransport}, construction of living bridges \cite{garnierBridges}, schooling \cite{strandburgVisual}, and flocking \cite{balleriniTopological}. There is no central planner in these biological systems; instead, interactions based on locally-available information leads to such coordination \cite{camazineBook}.

Efforts to describe the rules that determine whether two individuals in a group are permitted to interact (i.e., the network topology that underpins communications) has resulted in numerous models being proposed in the biological swarm literature. Three predominant models have been developed to describe the communication: the metric \cite{couzinLeadership}, the topological \cite{balleriniTopological}, and the visual models \cite{strandburgVisual}. The swarm's agents interact if they are within a critical distance of one another in the metric model; hence, this model is directly based on spatial proximity \cite{couzinLeadership}. Ballerini {\it et al.}'s topological model \cite{balleriniTopological} is similar in concept to the nearest neighbor rule $(k-NN)$ \cite{machineLearningBook}, in that, it requires individuals to interact with a fixed number of nearest individuals. The visual model is based on sensory capabilities, where an individual only interacts with those within its field of view \cite{strandburgVisual}. Other proposed communication models include those based on Delaunay triangulations \cite{ginelliVornoi,couzinComparison}, cognitive heuristics \cite{moussaidCrowd}, and selective attention \cite{lemassonMotion}.

Identifying the communication model that best describes a biological swarm is important to the science of collective behavior, as it provides  insight into how information diffuses in a swarm \cite{strandburgVisual}. The development of communication models is also important from a robotics perspective, and is described as ``one of the main challenges" in swarm robotics \cite{hsiSurvey}. Biologically inspired artificial swarms derive characteristics from their biological counterparts, and in addition to bypassing centralized control laws, there are other benefits to designing engineered systems inspired by social animals, such as scalability and robustness to individual agent failures \cite{bonabeauBook}. A survey \cite{hsiSurvey} of human-swarm interaction notes that despite inheriting beneficial characteristics from their counterparts in nature, an ill-conceived communication model can lead to undesirable consequences. The authors posit that erratic behavior resulting from a poorly designed communication model increases the workload of a human operator interacting with the swarm.

This manuscript's findings demonstrate that the choice of a communication model -- metric, topological, or visual -- is an important swarm design consideration, since the communication model has a significant impact on an artificial swarm's task performance. Six tasks were analyzed: {\it Search for Multiple Targets}, {\it Search for a Goal}, {\it Rally}, {\it Disperse}, {\it Avoid an Adversary}, and {\it Follow}, and a breadth of performance metrics were recorded to judge the artificial swarm's ability to conduct a task. No single communication model delivered the best performance across all the tasks. Further, agent and environmental parameters had meaningful interactions with the communication models in terms of task performance. The results imply that the performance of a deployed artificial swarm is amplified through a task-based selection of a communication model. In addition, the choice of a communication model can be fine-tuned, given environmental and agent parameters, such as the swarm's size. No prior research has conducted such an extensive analysis of the biologically inspired communication models within the context of artificial robotic swarm tasks.

An understanding of the appropriate communication model to task specification has the potential to make it easier for a human operator to monitor and supervise an artificial swarm. A communication model that improves the swarms' likelihood to complete a task will reduce the human's workload associated with monitoring the task. Understanding the exact implications on human interaction is beyond the scope of this manuscript. Rather, the manuscript's contribution focuses on factors related to the model-task pairings and the importance in their consideration for artificial swarm design.


\section{Related Work}
\label{sec:relatedwork}


The prior research comparing communication models can be classified based on research motivations: 1) Identify the model that accurately describes the network topology of a biological swarm (biology), 2) Understand the differences between the models from their system-theoretic properties (physics), and 3) Determine the manipulability of models in terms of human-swarm interaction (computer science).

The metric model is one of the earliest models developed to represent range-limited communication between biological swarm agents \cite{brederFish,aokiFish,reynoldsBoids}. This model is used as a benchmark for comparison testing relatively newer communication models \cite{strandburgVisual,balleriniTopological,couzinComparison}. The model is widely-used in the field of multi-robot systems, as well, due to its ability to capture sensor range constraints \cite{gaziPassino,jadbabaieTAC,mengConnectedness}. However, a field study of European starlings, {\it Sturnus vulgaris}, indicates that the swarm uses a topological, rather than a range limited model \cite{balleriniTopological}. Specifically, starlings coordinate with their nearest six to seven neighbors (topological distance). An artificial swarm, in response to a simulated predator, decomposes into fewer groups, and produces more cohesive swarms, when using the topological model compared to the metric model \cite{balleriniTopological}. Strandburg-Peshkin {\it et al.} \cite{strandburgVisual} introduced the visual model, and show that it best predicts how golden shiners, {\it Notemigonus crysoleucas}, behave in response to stimuli. The model's low clustering makes it fundamentally different from the metric and topological models, from a network-theoretic perspective.

Physics-based investigations of system-level properties of the topological and metric models found group orders \cite{barberisTopological,bouffanais3}, the probability of reaching a consensus \cite{bouffanaisConsensus}, rate of convergence of the consensus on agents' headings \cite{bouffanaisInfluence}, and the influence of the topological distance on a simulated swarm's ability to reach a consensus in the presence of uncertainty \cite{youngStarlings}. Spears {\it et al.} \cite{spearsPhysicomimetics} did not explicitly compare the three leading models in their ``physicomimetic" simulated swarms, but compared swarm behaviors designed to be analogous to molecules in solid, liquid, and gas formations. This work was motivated by an unevaluated hypothesis that each swarm type (solid, liquid or gas) is particularly better suited than the other two in performing certain tasks: a solid swarm is better at distributed sensing; liquid at obstacle avoidance; and gas at surveillance-like coverage.

Goodrich {\it et al.} \cite{goodrichTR} compare the topological and metric models in order to evaluate a human's ability to control an artificial swarm by manipulating a leader agent that influences other swarm agents. Reportedly, a human operator can more easily manipulate a swarm using the topological model, and in general, swarms that have low inter-agent influences \cite{goodrichInvariants}. Other studies compare network topologies: De la Croix and Egerstedt \cite{jpLeader}, for instance, report on the ease with which a human operator can control a swarm whose communication network can either be a line, cycle, acyclic, or a complete graph. A single leader was controlled using a joystick. Multiple, dynamically assigned leaders were analyzed in networks, where agents were guaranteed to be 1-, 2-, or 3-hops from a leader \cite{walkerLeaders}. The work can be interpreted as comparing select topological distances to a swarm leader.

This manuscript's evaluation is seemingly the first to compare biologically inspired communication models with respect to their performances over artificial swarm robotics tasks.


\section{Coordination Algorithms}
\label{sec:algorithms}


One of the major aspects in which artificial swarms differ from typical multi-robot systems is in their coordination design \cite{hsiSurvey,swarmReviewed}. The individual agents in multi-robot systems are generally capable of performing tasks on their own; for instance, consider the system of robots described by Matari{\'c} \cite{mataricRL} and Burgard {\it et al.} \cite{thrunExploration}. Such systems benefit from the coordination amongst members, but such a characteristic is not a system-level requirement when agents are planning their own actions. However, a swarm, by definition, consists of ``relatively incapable" units, and through simple interaction rules, a global system behavior {\it emerges} \cite{sahinRobotics} \cite{brambillaReview}.

Incapable swarm units are conceivably limited in their ability to execute intricate interaction rules. Therefore, the designed coordination algorithm defining the movement laws aims to remain simple. The agents, modeled as $2D$ self-propelled particles, are controlled through updates to the velocity heading \cite{vicsekSPP,couzinRadii,goodrichWhat}. The agents are indexed $1$ through $N$, where $N$ is the swarm's size. Each agent $i\in\{1,\ldots,N\}$ experiences a force at each time iteration, $t$:
\begin{equation}
\label{eq:force}
\mathbf{F}_{i}(t)=\mathbf{F}_{env,i}(t)+\mathbf{F}_{swarm,i}(t)+\mathbf{F}_{task,i}(t), 
\end{equation}
where, $\mathbf{F}_{env,i}(t)$, $\mathbf{F}_{swarm,i}(t)$, and $\mathbf{F}_{task,i}(t)$ are the forces due to the environmental factors, swarming, and the task at hand, respectively. Such a framework of accumulating forces to control a swarm (Equation (\ref{eq:force})) is utilized to analyze the effectiveness of providing haptic feedback to a human operator \cite{coFields,goodrichShape,kollingTwoTypes}. $\mathbf{F}_{env,i}(t)$ incorporates reactions to the environment, such as remaining within the bounds of the simulated world by ``bouncing off" walls and avoiding obstacles. $\mathbf{F}_{task,i}(t)$ depends on the task (Section \ref{sec:tasks}). This force is not designed to optimally solve the associated robotics task; rather, it is a simple task-related objective that contributes to the overall force acting on an agent. The reason for this design choice is to gain insight into what the overall swarm can achieve with little intelligence guiding the individual units.

The choice of a communication model prescribes an agent's {\it neighbor set}. A communication link from $i$ to agent $j\in\{1,\ldots,N\}$, where $i\neq j$, classifies agent $j$ as agent $i$'s neighbor. $\mathcal{N}_i(t)$ denotes the neighbor set of agent $i$, and it represents the collection of agent $i$'s neighbors at time $t$.

The metric model is parameterized by a single measure on distance, $d_{met}$ (i.e., the metric range). All agents within a distance $d_{met}$ from agent $i$ are its neighbors, as shown in Figure \ref{fig:coordination}(a). Due to the symmetric nature of the model, $j\in\mathcal{N}_i(t)$ implies that $i\in\mathcal{N}_j(t)$.

$\mathcal{N}_i(t)$, as assigned by the topological model, is the set containing the $n_{top}$ nearest agents from agent $i$, where $n_{top}$ is referred to as the topological distance. The topological distance of Zebrafish, {\it Danio rerio}, is between three and five \cite{porfiriTopological}, whereas the distance for starlings is approximately six to seven \cite{balleriniTopological}. Figure \ref{fig:coordination}(a) depicts a network with $n_{top}$ set to four.

An agent's visual sensing is defined by a range $d_{vis}$ and an angle $\pm\phi$ from its heading \cite{couzinRadii,goodrichWhat}, which is a geometric construction that can produce a blindspot (Figure \ref{fig:coordination}(a)). The visual model prescribes neighbors based on three factors. Agent $j$ is agent $i$'s neighbor, if the following conditions hold: 1) Agent $j$ is not in agent $i$'s blindspot, 2) The two agents are less than $d_{vis}$ apart (the visual range), and 3) There is a clear line-of-sight between the agents (which can be occluded by another agent or object in the environment) \cite{strandburgVisual}.

$\mathbf{F}_{swarm,i}(t)$, is thus developed using a two-step process. The first step assigns ``neighbors" to an agent. Then, agents {\it swarm} with their neighbors based on the widely-used repulsion--orientation--attraction scheme \cite{couzinComparison,aokiFish,reynoldsBoids,vicsekSPP,couzinRadii,huthWissel,swarmConsumption,starlingsProb}. At each time $t$, agent $i$ experiences:
\begin{equation}
\label{eq:swarm}
\mathbf{F}_{swarm,i}(t)=\mathbf{F}_{r,i}(t)+\mathbf{F}_{o,i}(t)+\mathbf{F}_{a,i}(t), 
\end{equation}
where, $\mathbf{F}_{r,i}(t)$ pushes agent $i$ away from neighbors within a distance $r_{r}$, $\mathbf{F}_{o,i}(t)$ aligns it with neighbors that are between a distance of $r_{r}$ and $r_{o}$, and $\mathbf{F}_{a,i}(t)$ pulls it toward neighbors that are between a distance of $r_{o}$ and $r_{a}$.


\section{Experimental Design}
\label{sec:experimentaldesign}


\subsection{Setup}

\begin{figure}
\centering
\subfloat[The focus agent (fill) is neighbors with agents 3, 5, and 6 when using the metric model. The agent's neighbors using the topological model with $n_{top}$ = 4 are agents 3, 4, 5, and 6. The neighbors using the visual model are agents 3, 4, and 6 (agent 2 is occluded by 3, and agent 5 is in the blindspot).]{\includegraphics[scale=1.00]{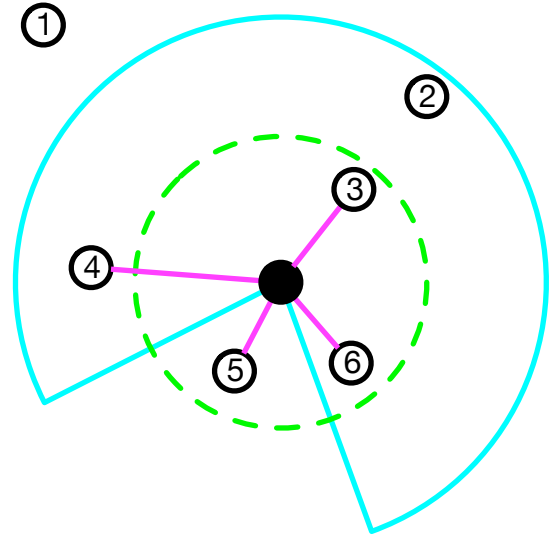}}
\hfill
\subfloat[The experiment's eight possible types of repulsion-orientation-attraction configurations \cite{cimaSwarms}. The inner-, middle, and outer-most zones represent the repulsion, orientation, and attraction zones, respectively, centered at the agent's position.]{\includegraphics[scale=0.90]{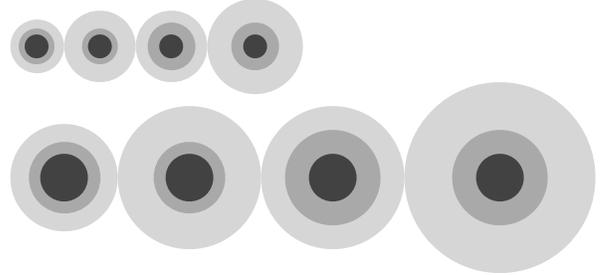}}
\caption{The choice of a communication model -- metric (green, dashed circle with radius $d_{met}$), topological ($n_{top}$ magenta lines), or visual (blue sector with radius $d_{vis}$ and $\pm \phi$ from heading) -- prescribes neighbors to the agents. Subsequently, agents interact with their neighbors by following a repulsion-orientation-attraction scheme.}
\label{fig:coordination}
\end{figure}

The {\tt Processing} development environment\footnote{https://processing.org/} was used to conduct the experiments. Across all tasks, the communication model was the experiment's primary factor with additional factors being the number of agents ($N$), and the radii of repulsion ($r_{r}$), orientation ($r_{o}$), and attraction ($r_{a}$). The communication model was set to either metric, topological, or visual. $N$ had three levels: 50, 100, and 200 agents. $r_r$ was set to either 10 or 20 pixels. $r_o$ was either 1.5$\times r_r$ or 2.0$\times r_r$. Similarly, $r_a$ was either 1.5$\times r_o$ or 2.0$\times r_o$. The resulting radii configurations are shown in Figure \ref{fig:coordination}(b). Some tasks utilized additional factors, which are specified, along with their explored levels, in Section (\ref{sec:tasks}). The experiment combined the primary factor with each of the additional factors, producing pairwise combinations that offered a more comprehensive analysis of the effect of the communication models.

The biological swarm literature guides the parameter value selection of $d_{met}$, $n_{top}$, $d_{vis}$, and $\phi$. The metric range, $d_{met}$, was set to $r_{a}$, following Couzin {\it et al.} \cite{couzinLeadership}. The topological distance, $n_{top}\in\{5, 6, 7, 8\}$, permitted variability, while remaining close to what was observed in nature \cite{balleriniTopological}. However, only $n_{top} = 7$ is reported for the topological model, without loss of generality. No difference in performance was found between the different levels of $n_{top}$ across all the tasks, and this characteristic of the topological model can be attributed to the existence of a critical topological distance $n_{top}^\star$, beyond which the swarm's performance does not vary \cite{irosSwarms}. The visual range, $d_{vis}$, was assigned to half the size of the diagonal of the world, with $\phi=2\pi/3$ radians \cite{strandburgVisual,couzinRadii}. This attempt to derive values from what is reported in the biological swarm literature, adheres to the ``descriptive agenda" of multi-agent learning \cite{shammaQuestion, shohamQuestion}, where the goal is to model an underlying phenomenon from the social sciences.

$\mathbf{F}_{env,i}(t)$ is responsible for reflecting agent $i$ off walls \cite{natureOfCodeBook} by adding an offset to the current heading near obstacles. Certainly, obstacles can be avoided more intelligently (using collision cones \cite{kristiCones} or barrier certificates \cite{danielRobotarium}, for instance), but the reason to not employ such techniques is to allow the models to drive the coordination without the help of sophisticated maneuvers.

\begin{figure*}
\centering
\subfloat[Visual, t = 170]{\includegraphics[scale=0.18]{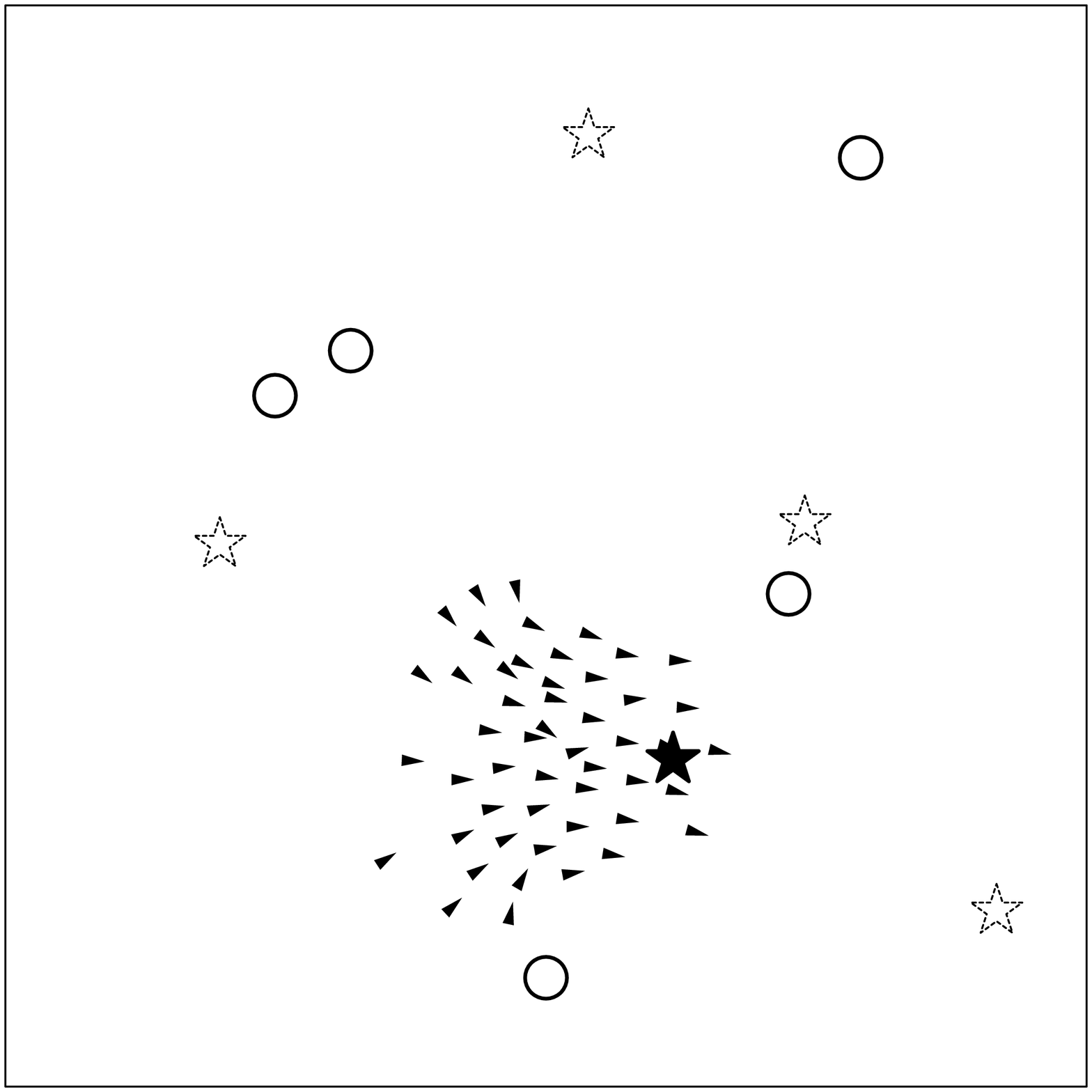}}
\subfloat[Visual, t = 585]{\includegraphics[scale=0.18]{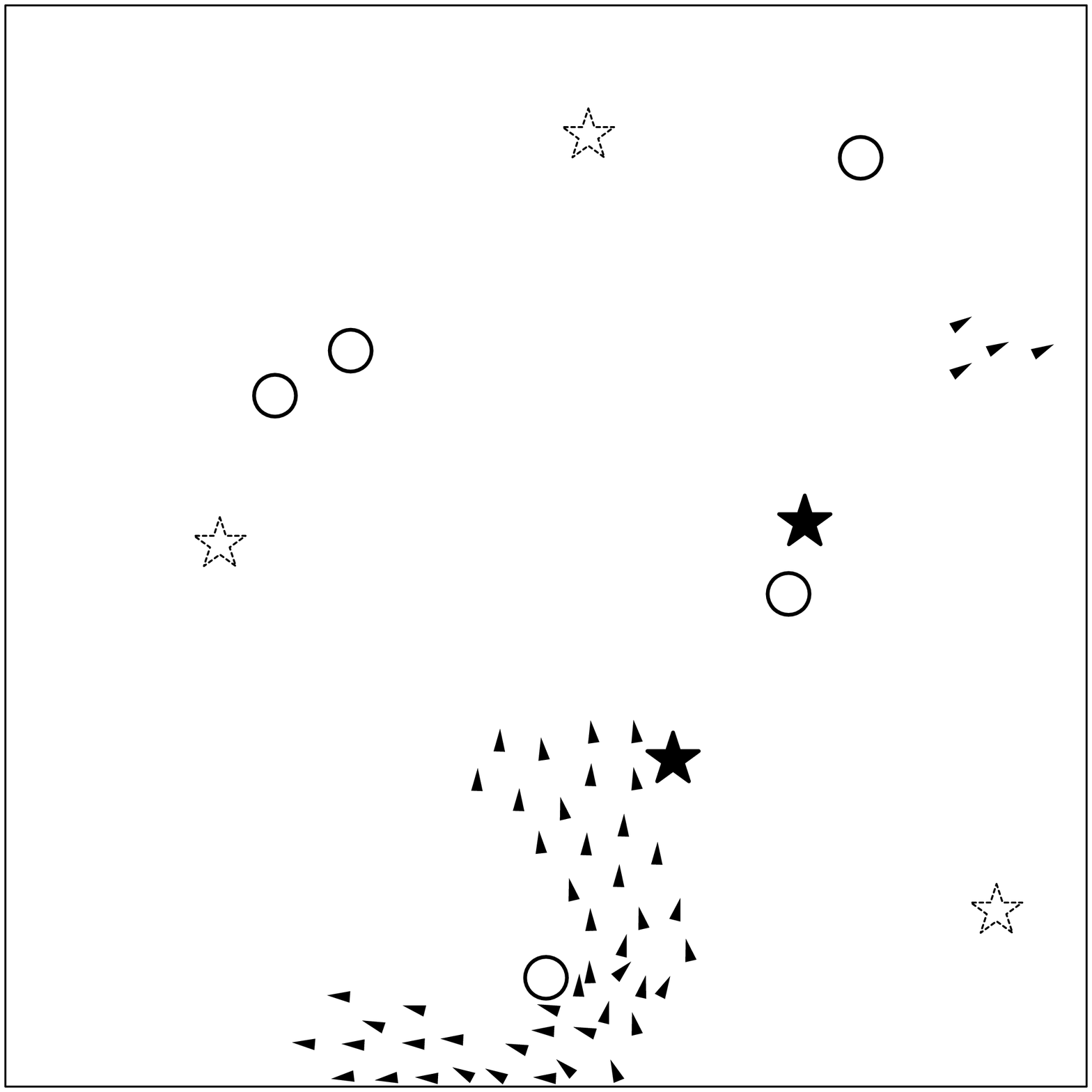}}
\subfloat[Visual, t = 990]{\includegraphics[scale=0.18]{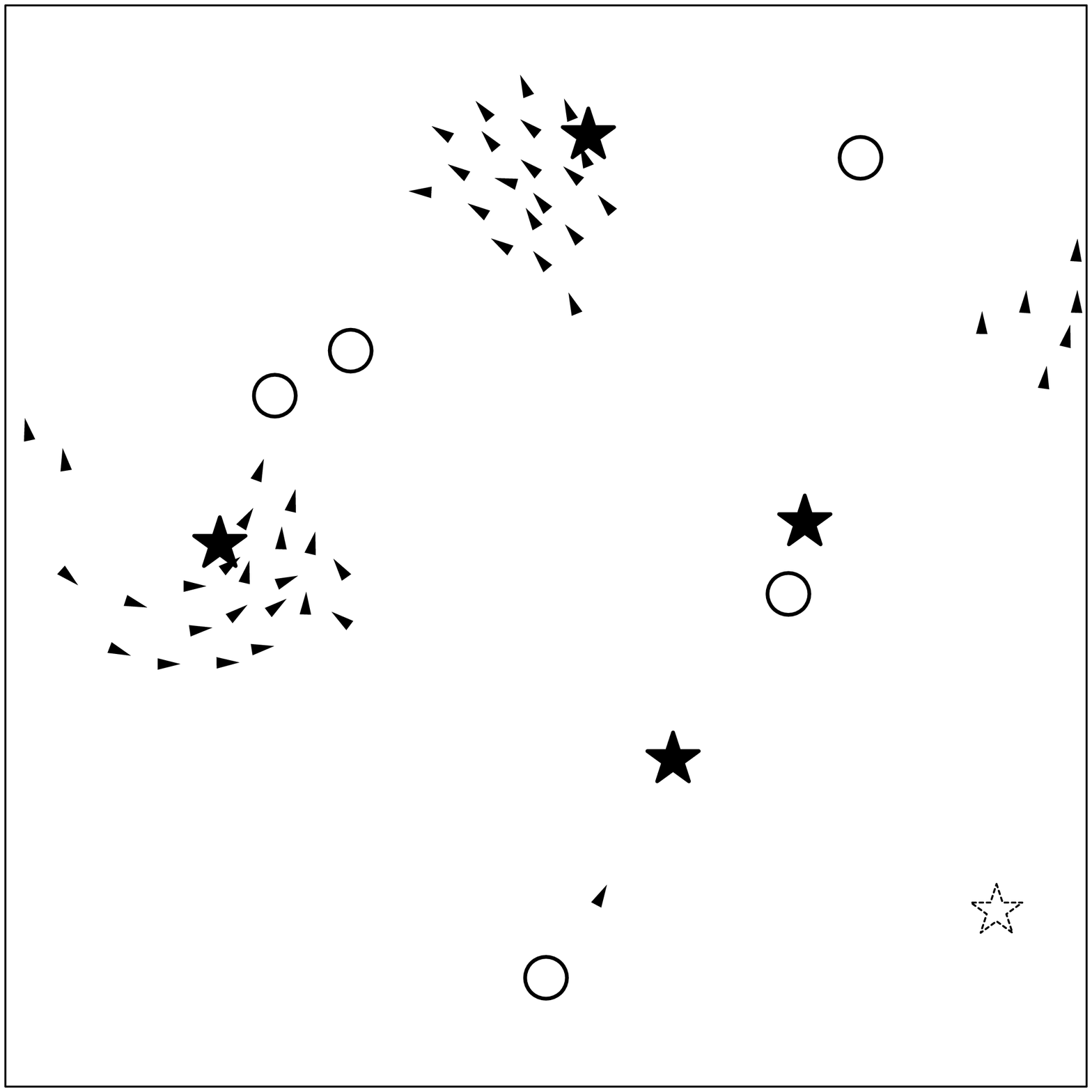}}
\hfill
\subfloat[Topological, t = 35]{\includegraphics[scale=0.18]{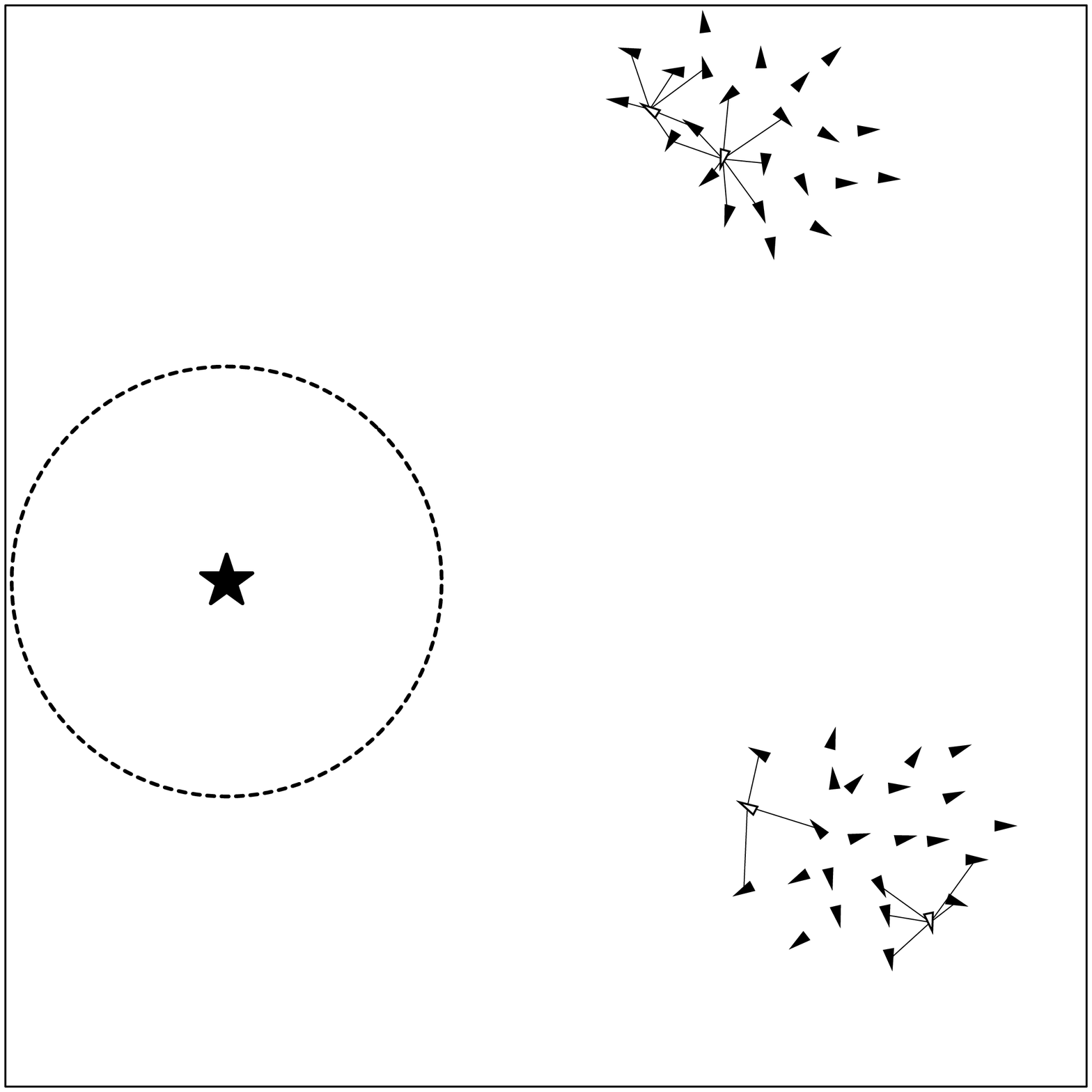}}
\subfloat[Topological, t = 130]{\includegraphics[scale=0.18]{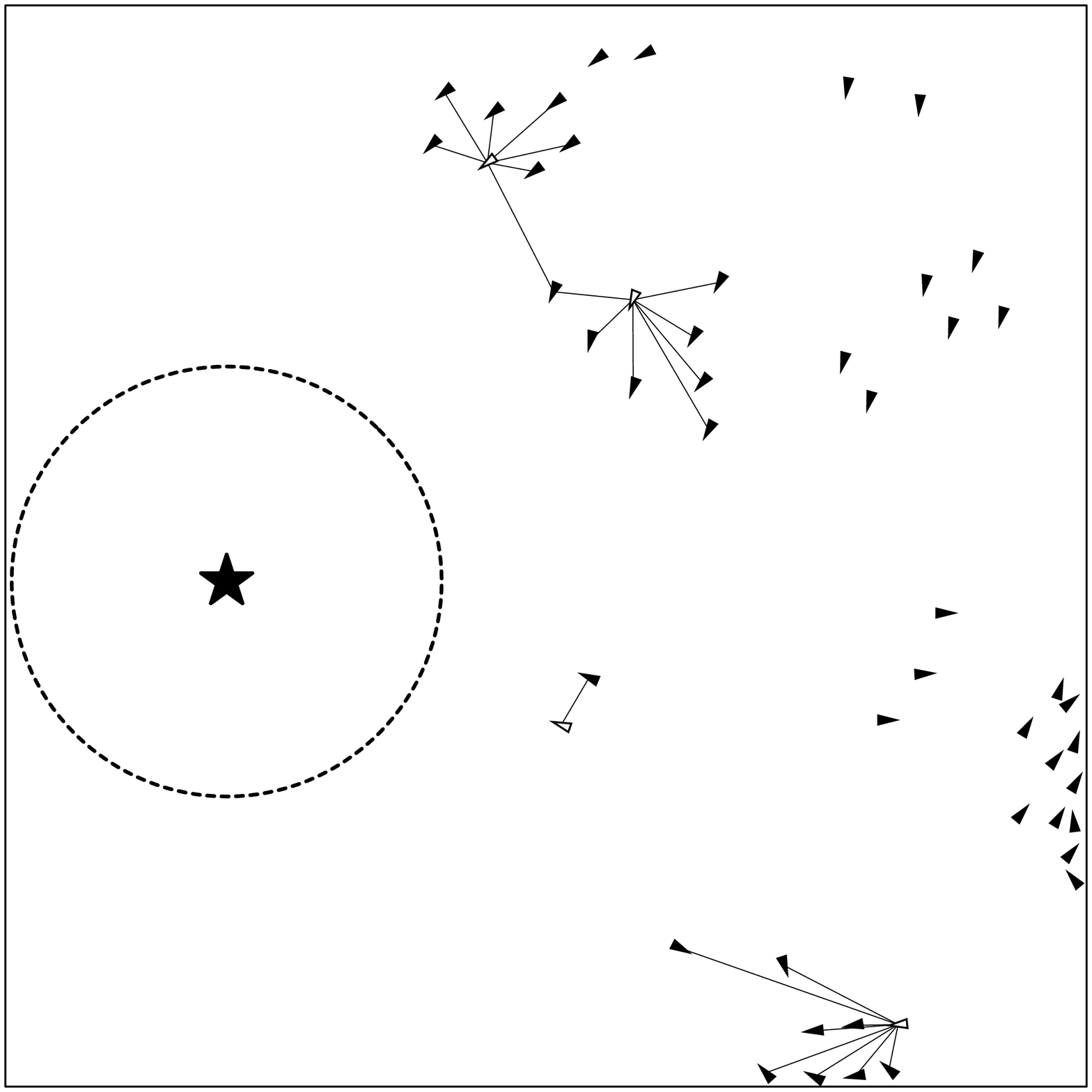}}
\subfloat[Topological, t = 669]{\includegraphics[scale=0.18]{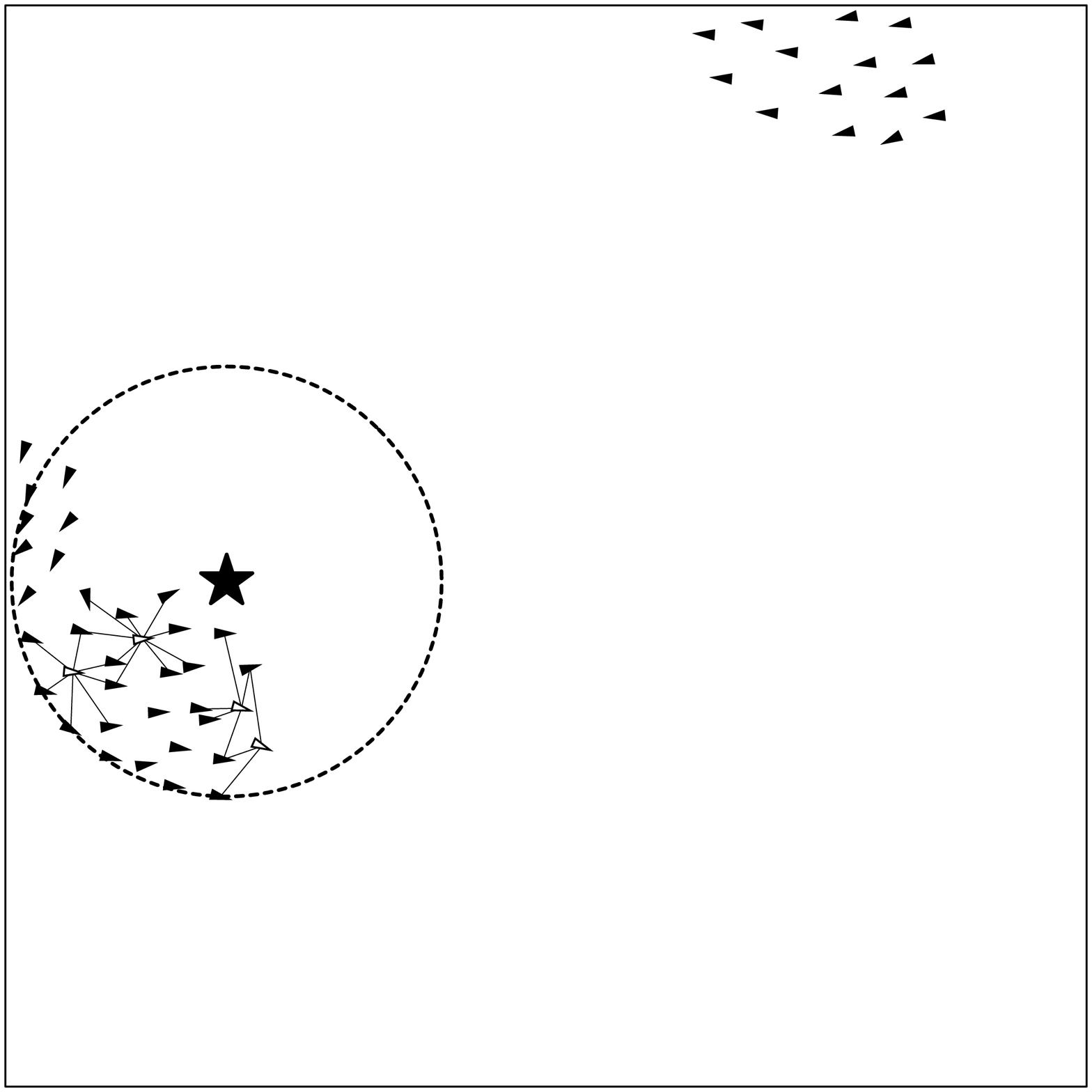}}
\hfill
\subfloat[Metric, t = 199]{\includegraphics[scale=0.18]{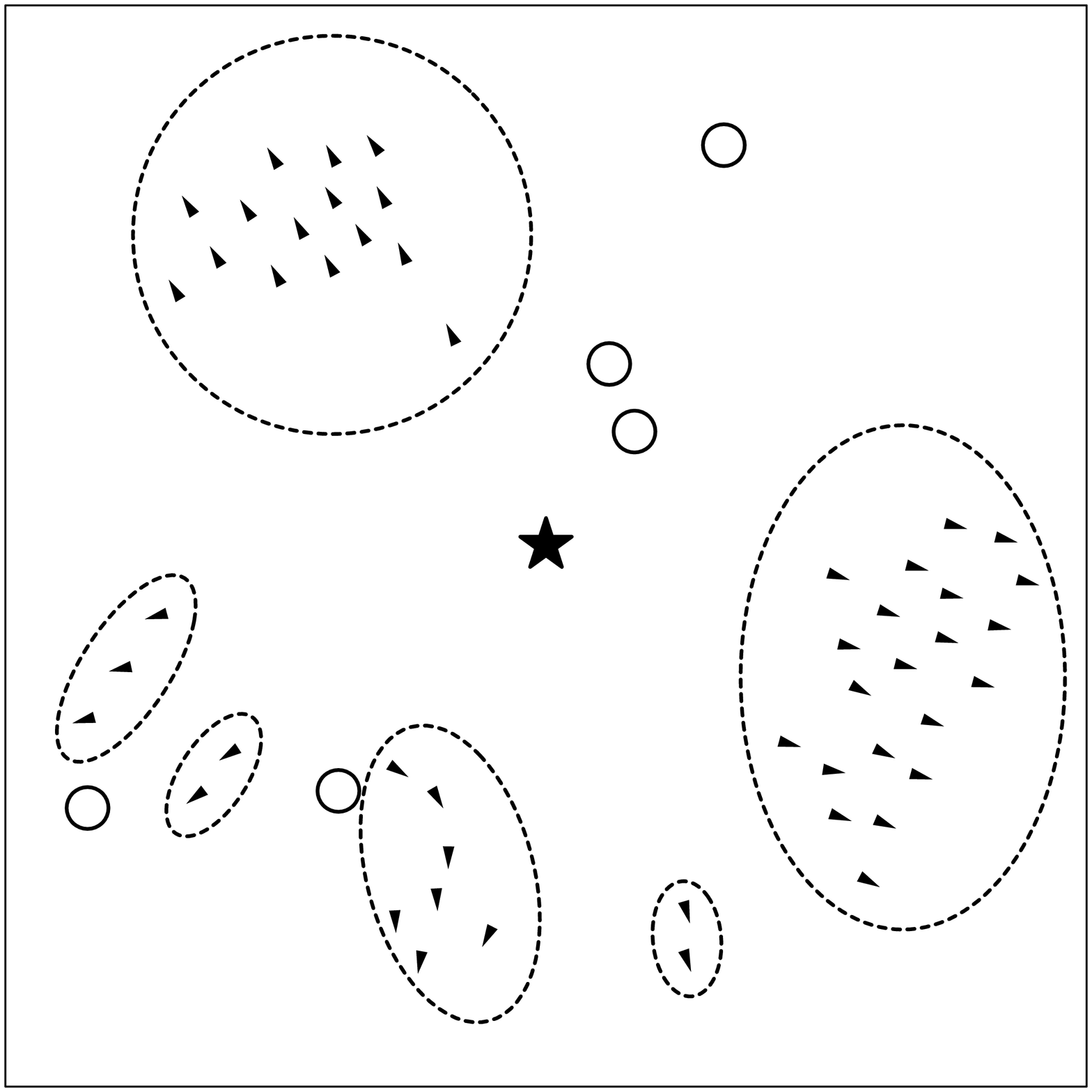}}
\subfloat[Topological, t = 199]{\includegraphics[scale=0.18]{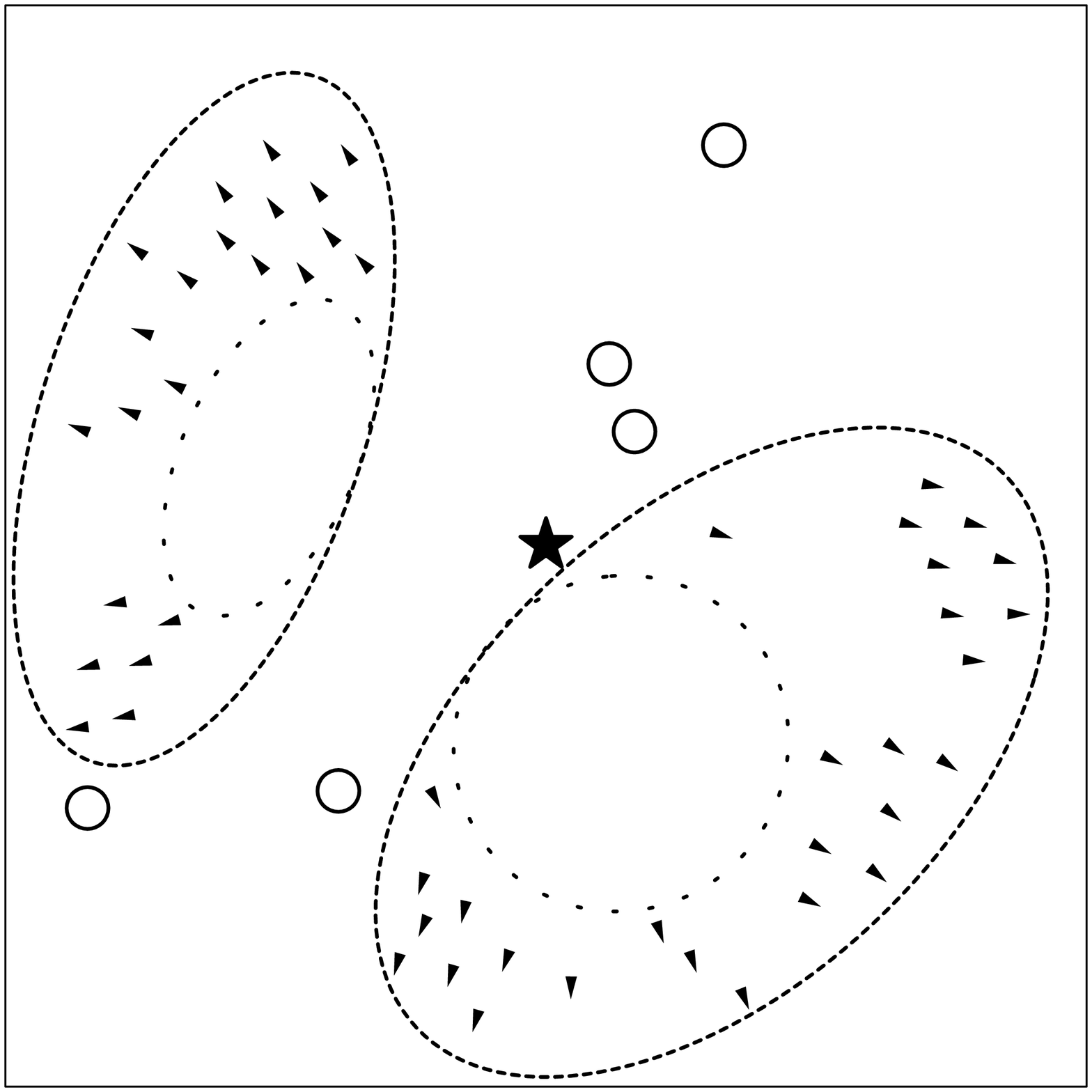}}
\subfloat[Visual, t = 199]{\includegraphics[scale=0.18]{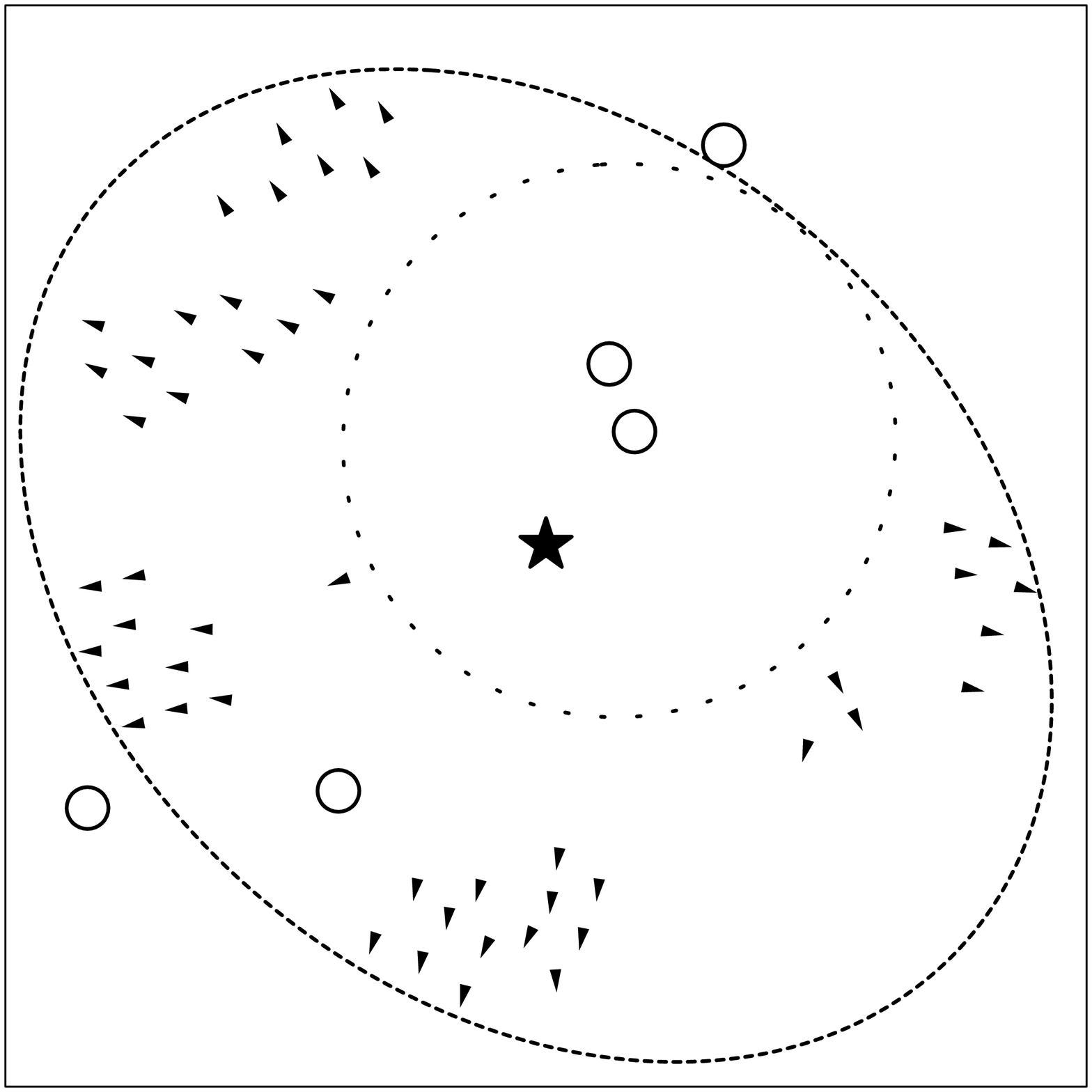}}
\hfill
\subfloat[Metric, t = 181]{\includegraphics[scale=0.18]{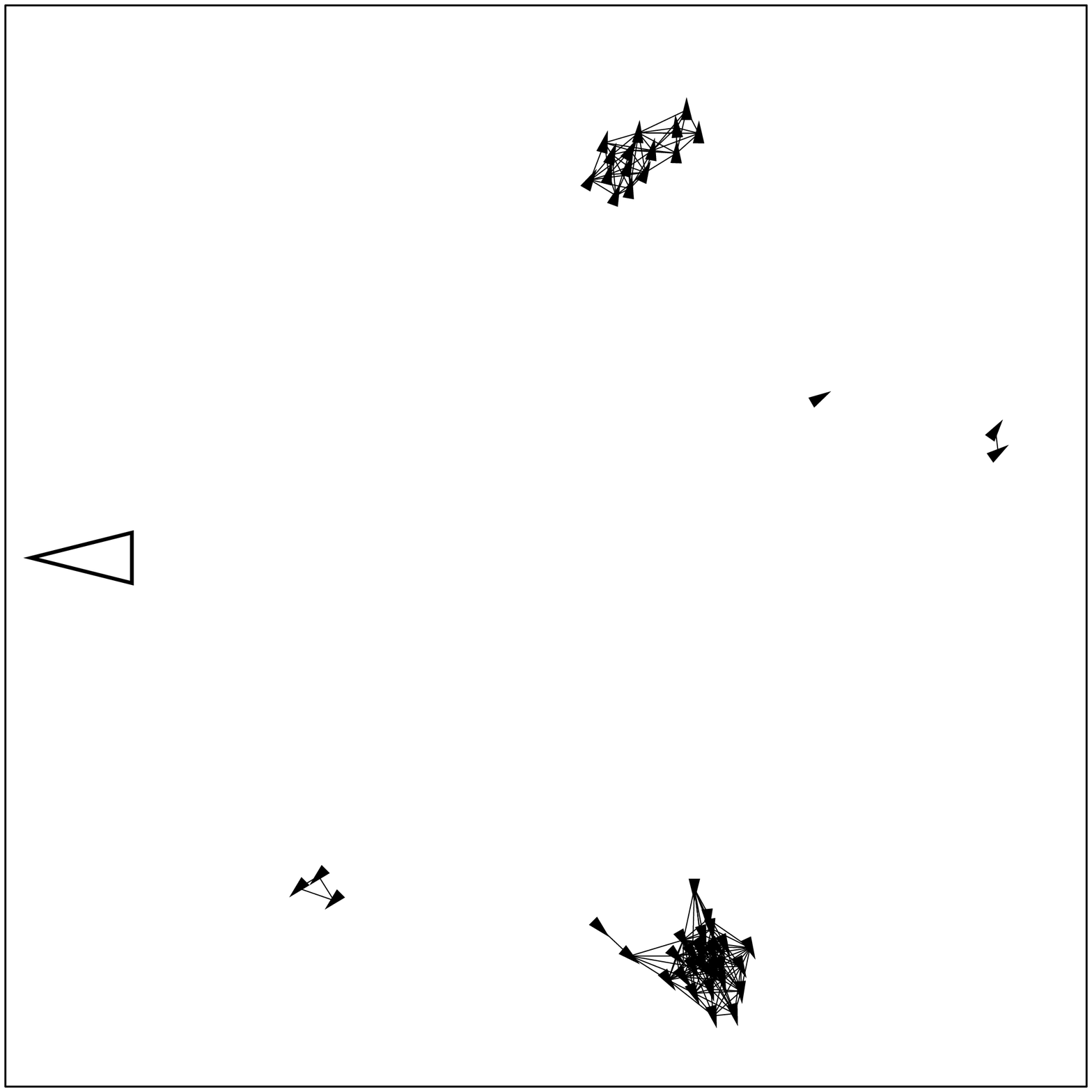}}
\subfloat[Topological, t = 187]{\includegraphics[scale=0.18]{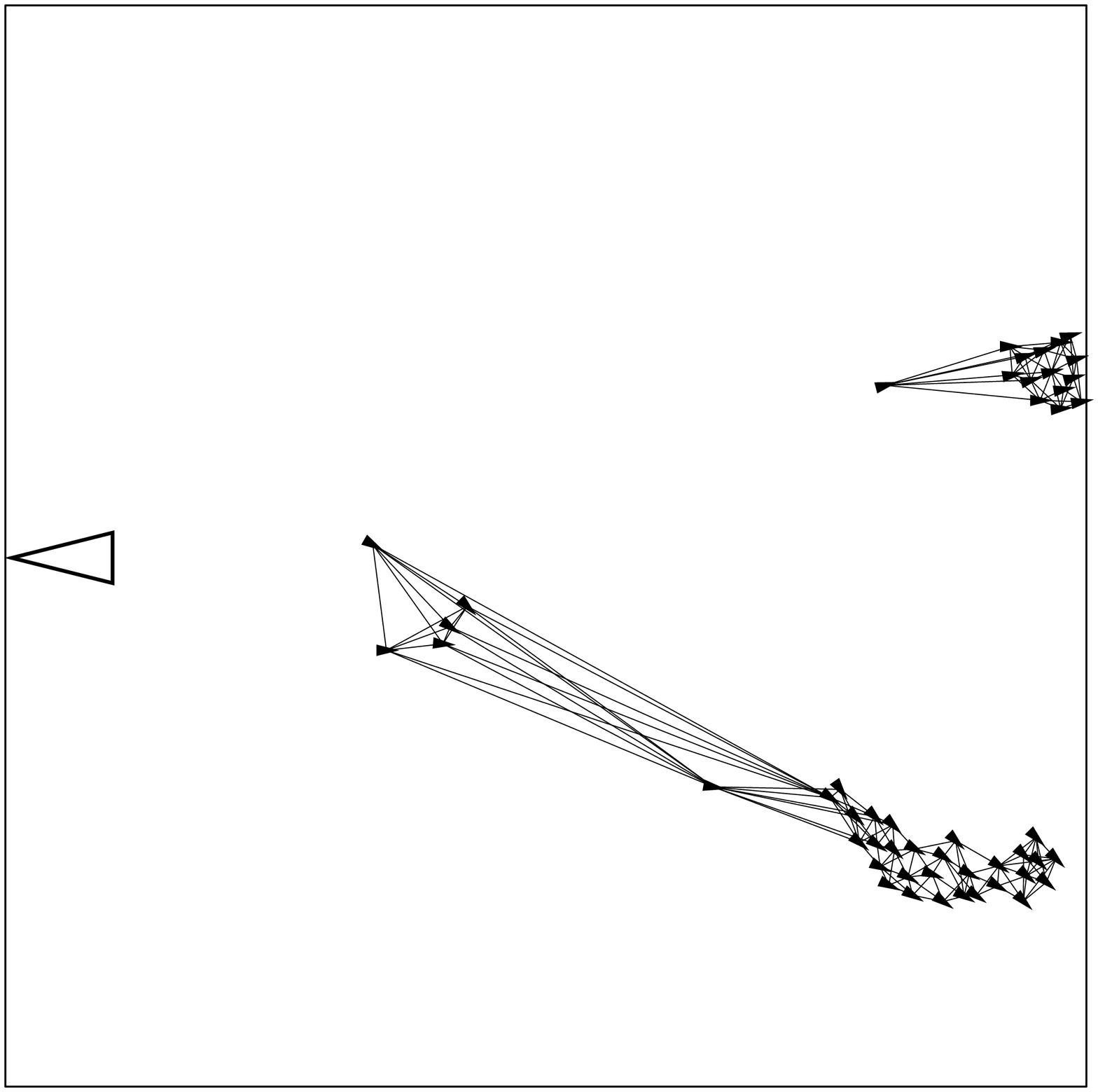}}
\subfloat[Visual, t = 183]{\includegraphics[scale=0.18]{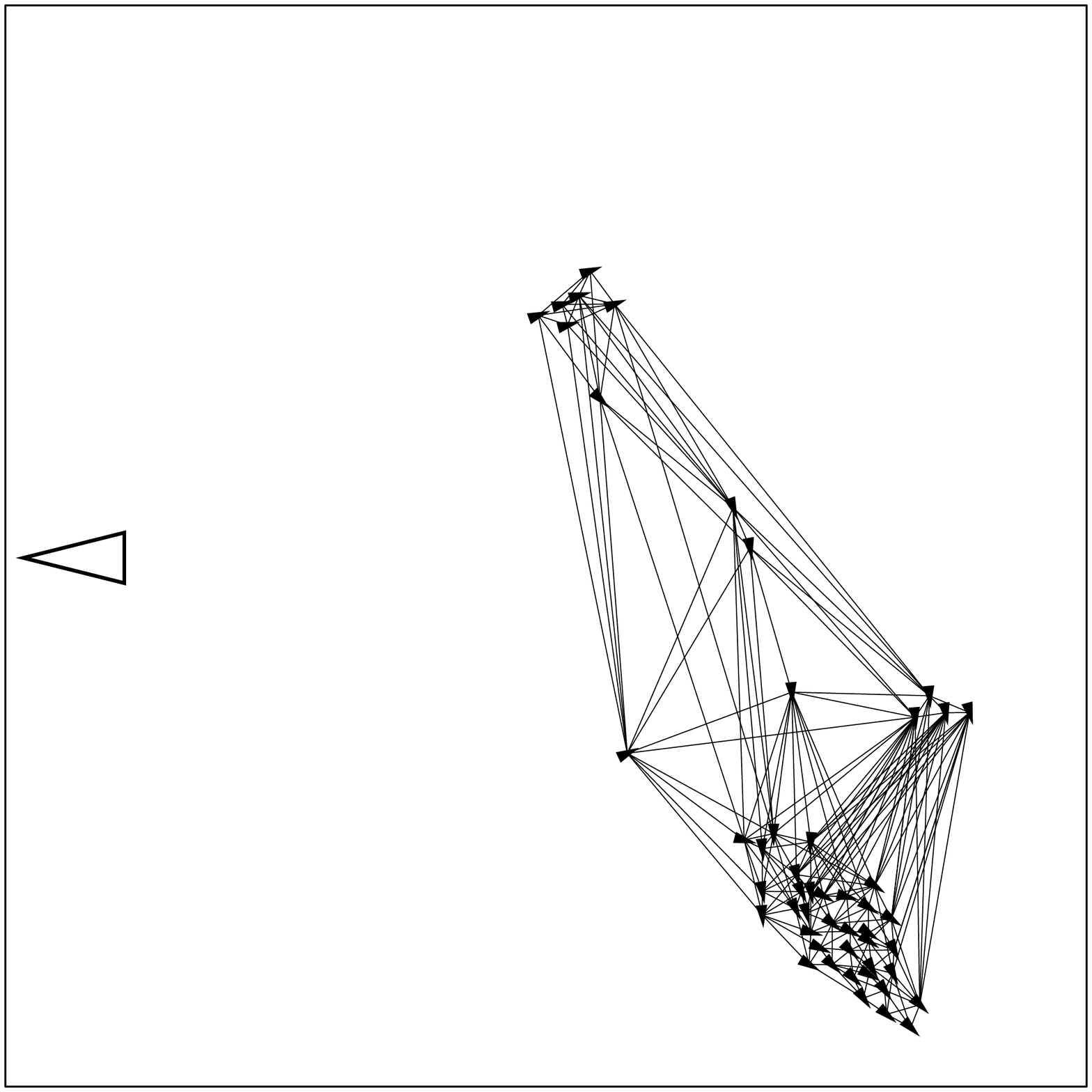}}
\hfill
\subfloat[Visual, t = 1212]{\includegraphics[scale=0.18]{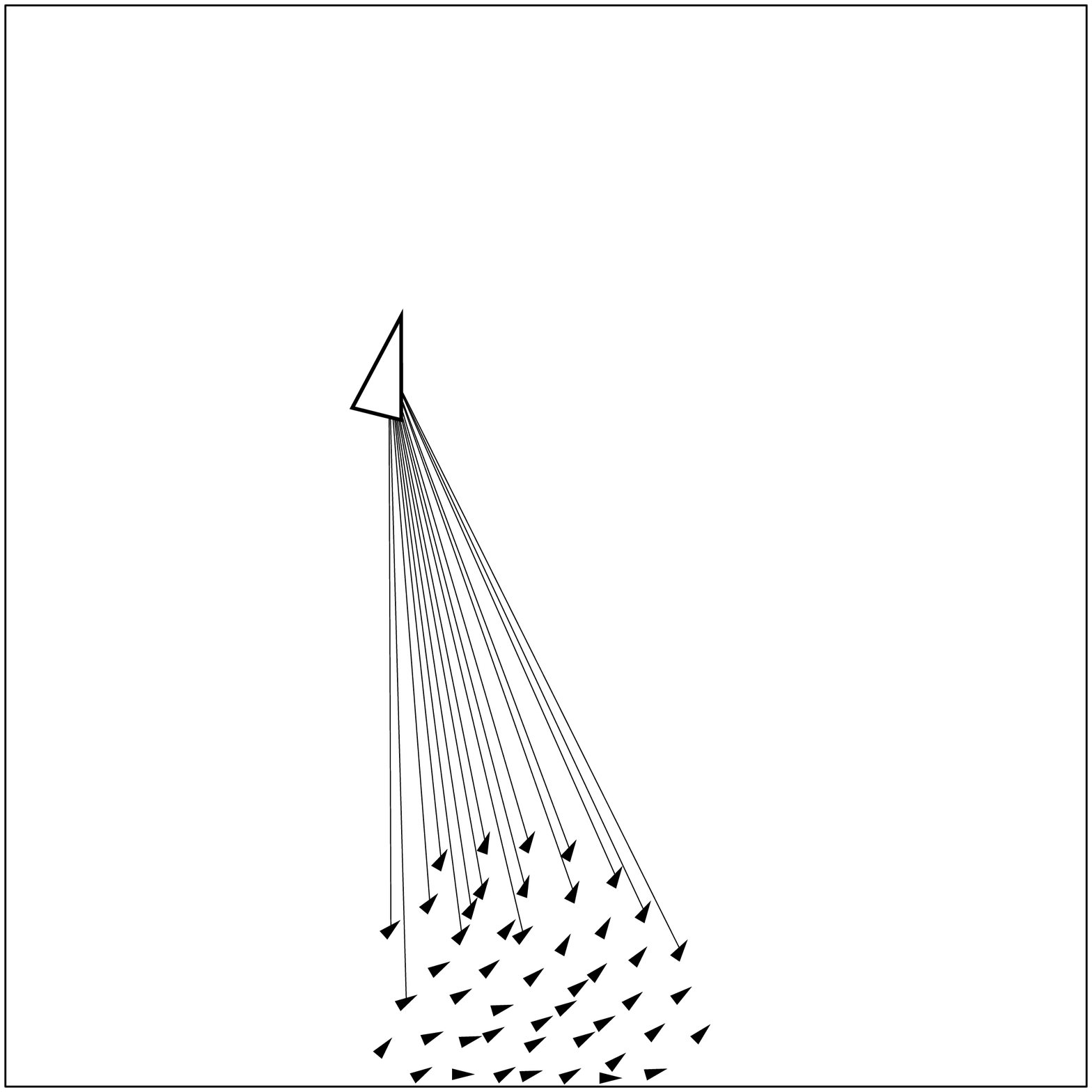}}
\subfloat[Visual, t = 1393]{\includegraphics[scale=0.18]{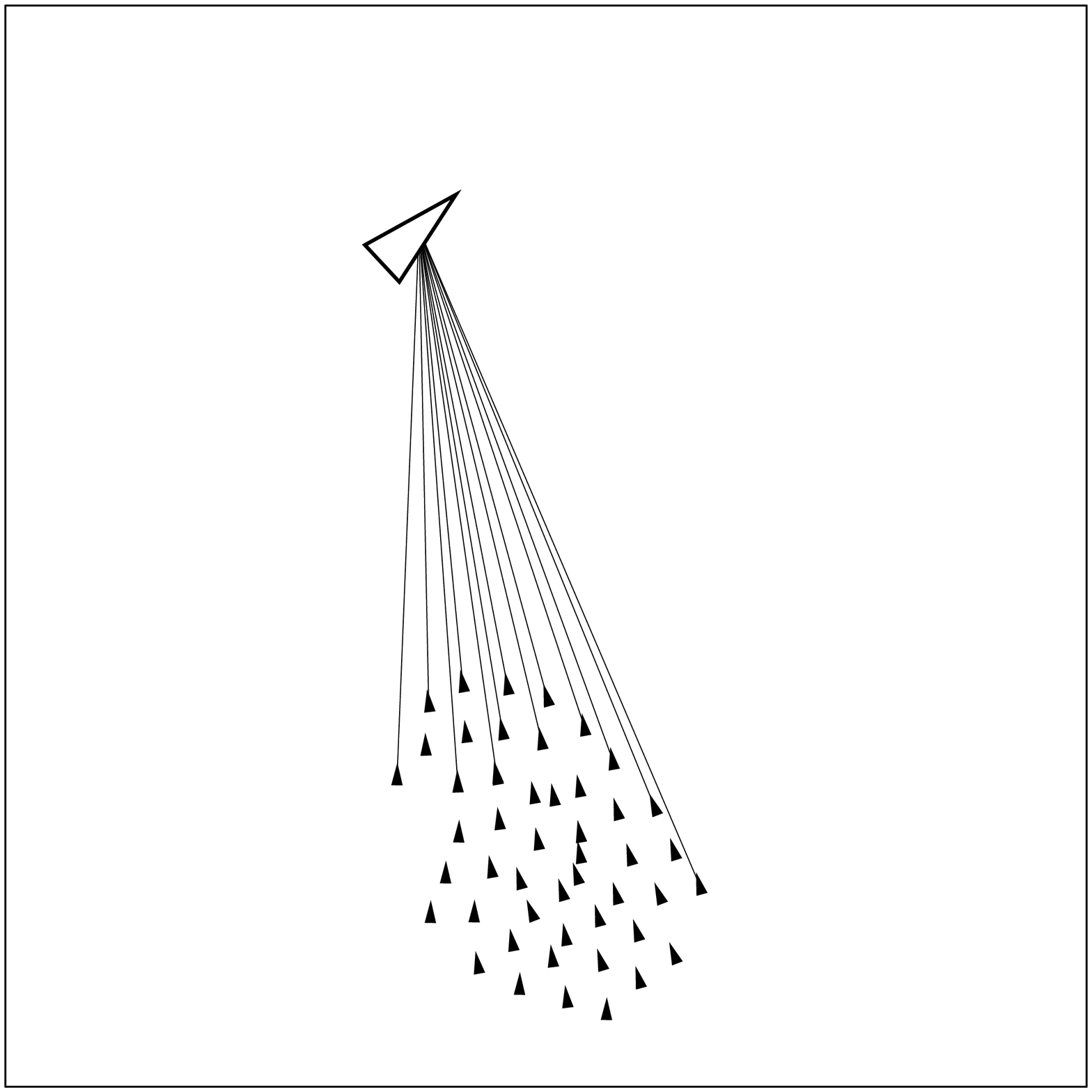}}
\subfloat[Visual, t = 1830]{\includegraphics[scale=0.18]{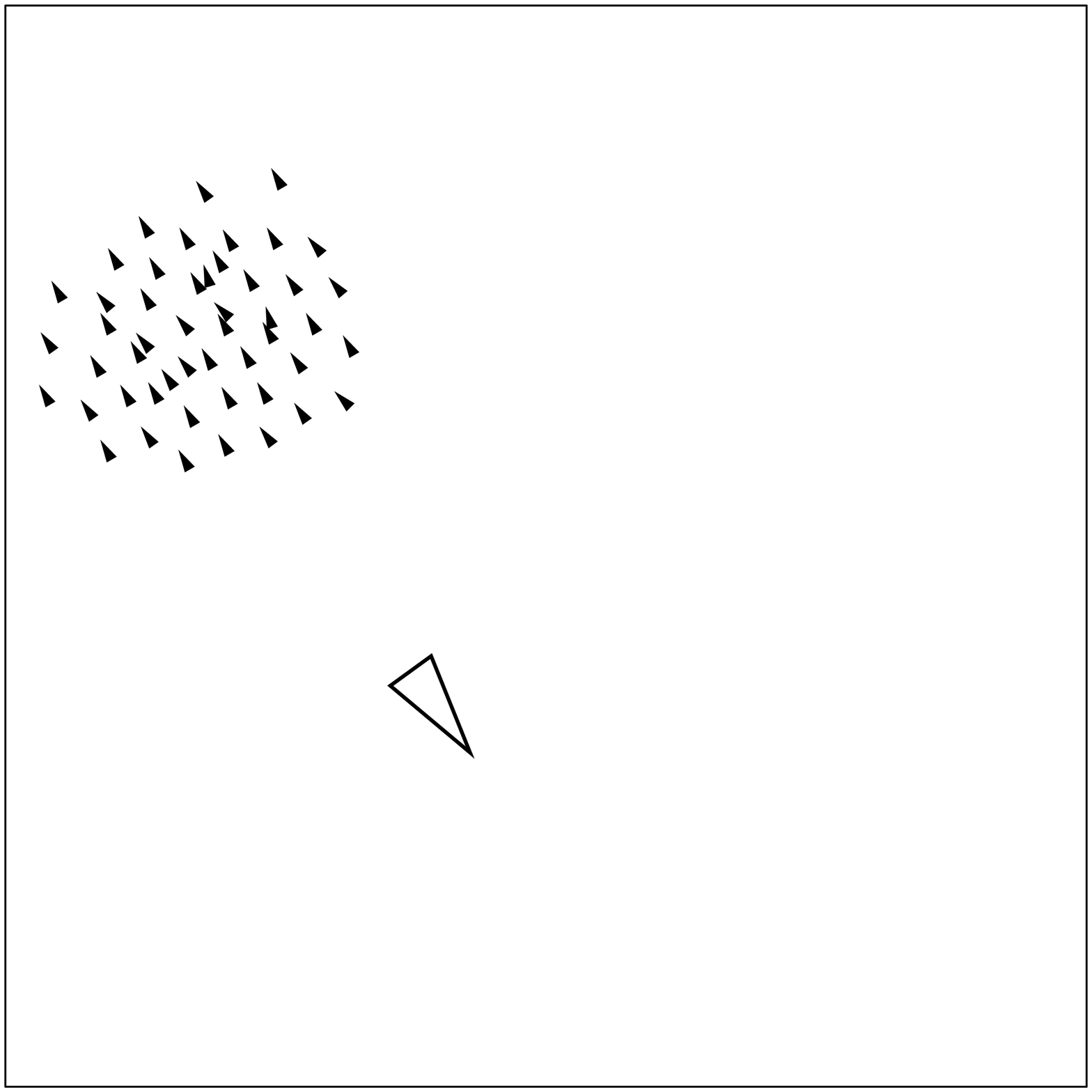}}
\caption{The predator/leader and obstacles are represented by enlarged triangles and circles, respectively. (a)-(c) {\it Search}: stars denote targets (fill indicates discovered). (d)-(f) {\it Rally}: links between informed (no fill) and uninformed (fill) agents are shown. (g)-(i) {\it Disperse}: links are contained within the donut shapes of the topological and visual trials. (j)-(l) {\it Avoid}: links between swarm agents are shown \cite{cimaSwarms}. (m)-(o) {\it Follow}: links are shown between agents and the leader.}
\label{fig:tasks}
\end{figure*}

\subsection{Tasks}
\label{sec:tasks}

\subsubsection{Search for Multiple Targets}

The artificial swarm's objective was to discover targets (stars in Figures \ref{fig:tasks}(a)-(c)). $\mathbf{F}_{task,i}(t)$ was set to $\emptyset$; hence, there was no force requiring agents to search, let alone do so intelligently. This formulation investigated achievement through swarming alone, contained in an area, and while avoiding obstacles.

\subsubsection{Search for a Goal}

This task included a single goal (star in Figures \ref{fig:tasks}(d)-(f)) that the swarm was required to locate. Once an agent located the goal, it communicated the location to its neighbors. $\mathbf{F}_{task,i}(t)$ was enabled when an agent located the goal area, which acted as an attractor. Within the framework of Equation (\ref{eq:force}), agents aware of the goal's location updated their headings by weighing the desire to travel to the goal and the desire to follow the interaction rules \cite{couzinLeadership,goodrichTR,ecaiSwarms}.

\subsubsection{Rally}

The objective was similar to the {\it Search for a Goal} task's objective, except informed agents were included that were aware of the goal location (i.e., the rally point). Moreover, informed agents did not communicate this location to their neighbors. Each informed agent balanced its desire to abide by the swarming forces with a desire to move towards the rally point, similar to $\mathbf{F}_{task,i}(t)$ described in the prior task. The percentage of informed agents ($p_i$) was 8$\%$, 16$\%$, or 24$\%$ of $N$. (A small fraction of the swarm acting as anonymous leaders has been shown to alter the group's direction \cite{couzinLeadership}.) Agents were initialized into starting groups ($g$) of 1, 2, or 4.

\subsubsection{Disperse}

This task required agents to distribute themselves in the environment. Agents began the task placed around the center of the simulation environment (star in Figures \ref{fig:tasks}(g)-(i)). Each agent experienced a dispersing force, $\mathbf{F}_{task,i}(t)$, modeled by exerting a constant radial force away from the center of the environment. The strength of the dispersal force ($s$), was set to 45$\%$, 90$\%$, and 135$\%$ of the swarming force.

\subsubsection{Avoid an Adversary}

The swarm avoided a predator-like agent, which was modeled with $\mathbf{F}_{task,i}(t)$ being a repulsive force exerted by the adversary \cite{balleriniTopological} on the swarm agents $r_{a}$ away. The swarm was initially aligned facing the predator. The predator (moving in a predefined path) was the same size as the agents (enlarged in Figures \ref{fig:tasks}(j)-(l)) and occluded the visual communication between agents.

\subsubsection{Follow}

The swarm followed a single, leader-like agent. $\mathbf{F}_{task, i}$ was modeled as an attractive force to the leader when the leader was an agent's neighbor. The leader was the same size as the swarm agents (enlarged in Figures \ref{fig:tasks}(m)-(o)), moved at the same speed, and randomly navigated the world.

\subsection{Trials}

A trial was defined as a single simulation run for a given selection of factors. Twenty-five trials for each parameter selection were completed. The total number of trials per task is summarized in Table \ref{table:trials}. The {\it Search for Multiple Targets} task, for instance, had 5,400 metric, 5,400 visual, and 21,600 topological trials (due to the four levels of $n_{top}$).

\begin{table}[h]
\caption{Trials and Iterations by Task.}
\label{table:trials}       
\begin{tabular}{lcrr}
\hline\noalign{\smallskip}
Task & Factors per model & Trials & Iterations per run\\
\noalign{\smallskip}\hline\noalign{\smallskip}
Targets & ($N$, $N_{o}$, $N_{t}$, $r_{r}$, $r_{o}$, $r_{a}$) & 32400 & $1000$ \\
Goal & ($N$, $N_{o}$, $r_{r}$, $r_{o}$, $r_{a}$) & 10800 & 1000\\
Rally & ($N$, $p_i$, $g$, $r_{r}$, $r_{o}$, $r_{a}$) & 32400 & 750\\
Disperse & ($N$, $N_{o}$, $s$, $r_{r}$, $r_{o}$, $r_{a}$) & 32400 & 200\\
Avoid & ($N$, $r_{r}$, $r_{o}$, $r_{a}$) & 3600 & 200\\
Follow & ($N$, $r_{r}$, $r_{o}$, $r_{a}$) & 3600 & 2000\\
\noalign{\smallskip}\hline
\end{tabular}
\end{table}

\subsection{Metrics}

The swarm's performance was measured through the consideration of an array of metrics, not all of which are reported. This manuscript focuses specifically on the metrics intended to provide evidence that artificial swarm design needs to consider the communication model and task pairing in order to optimize the overall swarm performance. Still, a set of metrics emerged that remained relevant across the different tasks (see Table \ref{table:metrics}). The analysis was constrained to the main effects of the communication model and the simple interactions between model and the additional factors.

\begin{table}[h]
\caption{Unrecorded ($\times$), recorded ($\circ$), and reported ($\bullet$) metrics.}
\label{table:metrics}       
\begin{tabular}{l|cccccc}
\hline\noalign{\smallskip}
Metric &Target&Goal&Rally&Disperse&Avoid&Follow\\
\noalign{\smallskip}\hline\noalign{\smallskip}
$NCC$&$\bullet$&$\times$&$\circ$&$\bullet$&$\bullet$&$\circ$\\
$PF$&$\bullet$&$\times$&$\times$&$\times$&$\times$&$\times$\\
$L$&$\times$&$\bullet$&$\times$&$\times$&$\times$&$\times$\\
$SCC$&$\circ$&$\bullet$&$\circ$&$\circ$&$\circ$&$\circ$\\
$PR$&$\times$&$\bullet$&$\bullet$&$\times$&$\times$&$\times$\\
$DINF$&$\times$&$\times$&$\bullet$&$\times$&$\times$&$\times$\\
$D$&$\circ$&$\circ$&$\circ$&$\bullet$&$\bullet$&$\circ$\\
$PIC$&$\circ$&$\circ$&$\circ$&$\bullet$&$\bullet$&$\circ$\\
$ASTK$&$\times$&$\times$&$\times$&$\times$&$\times$&$\bullet$\\
$SSTK$&$\times$&$\times$&$\times$&$\times$&$\times$&$\bullet$\\
$INF$&$\times$&$\times$&$\times$&$\times$&$\times$&$\bullet$\\
\noalign{\smallskip}\hline
\end{tabular}
\end{table}

The {\it percent found} ($PF$) measured the number of targets that have been discovered in the area. A target's classification was irreversibly changed from ``undiscovered" to ``discovered" once an agent was within 10 pixels from the target's location.

The {\it number of connected components} ($NCC$) was reported as an average over a trial's duration. A connected component is defined as the largest collection of agents in which any two agents are either connected directly by a communication link or indirectly via neighbors \cite{networksBook}.

The {\it percent reached} ($PR$) determined the fraction of the swarm that reached the goal (50 pixels around goal's center).

The {\it latency} ($L$) represented the total iterations required for the swarm to transition from a state where at least one agent knew the goal's location to all agents being aware. Degenerate cases set the latency to the trial's duration.

The {\it swarm clustering coefficient} ($SCC$) was the average clustering coefficient over the swarm. The clustering coefficient in networks is the fraction of pairs of an agent's neighbors that are neighbors with each other \cite{networksBook}. The asymmetric nature of links that resulted from the topological and visual models were ignored, following Strandburg-Peshkin {\it et al.}'s \cite{strandburgVisual} treatment of directed links, when comparing different communication models for fish data.

{\it Dispersion} ($D$) measured the percentage increase of the average agent--agent distance from the start to the end of a trial. This distance was one of the factors identified by Parrish {et al.} \cite{parrishEmergent} to characterize the emergent properties of fish.

The {\it percent isolated components} ($I$) represented the fraction of the swarm that had no neighbors.

{\it Direct influence} ($DINF$) was the fraction of the swarm directly connected to an informed agent.

{\it Influence} ($INF$) was the fraction of the swarm that directly or indirectly followed the leader, at least once.

{\it Agent stickiness} ($ASTK$) represented the number of iterations an agent followed the leader, averaged over the swarm.

The {\it Swarm stickiness} ($SSTK$) was the number of iterations during which at least one agent was following the leader.


\section{Results}
\label{sec:results}



\subsection{Search for Multiple Targets}
\label{sec:targets}

\begin{figure*}
\centering
\subfloat[{\it Search for Multiple Targets}]{\includegraphics[trim=54 3 15 30,clip,scale=0.50, angle=270]{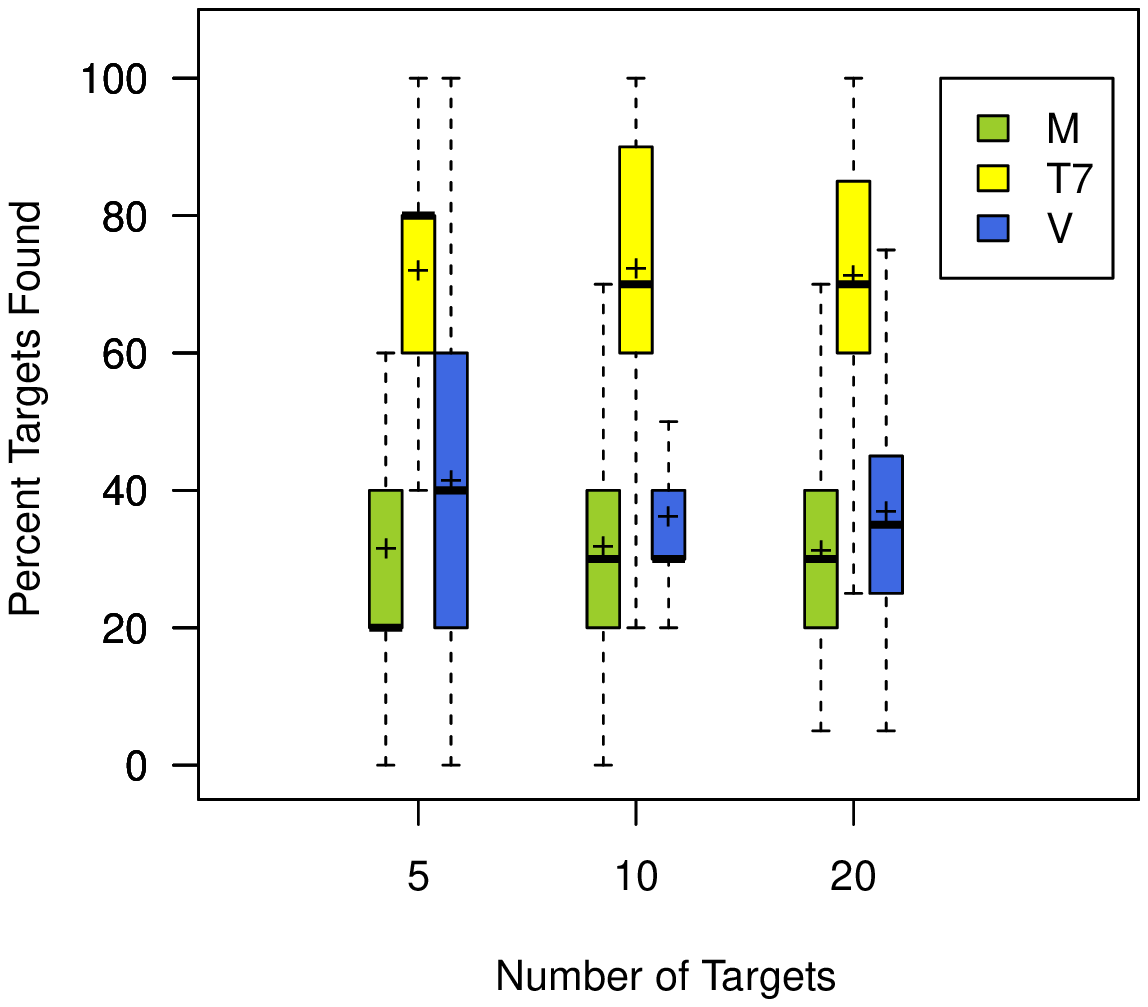}}
\hfill
\subfloat[{\it Search for Multiple Targets}]{\includegraphics[trim=54 3 15 30,clip,scale=0.50, angle=270]{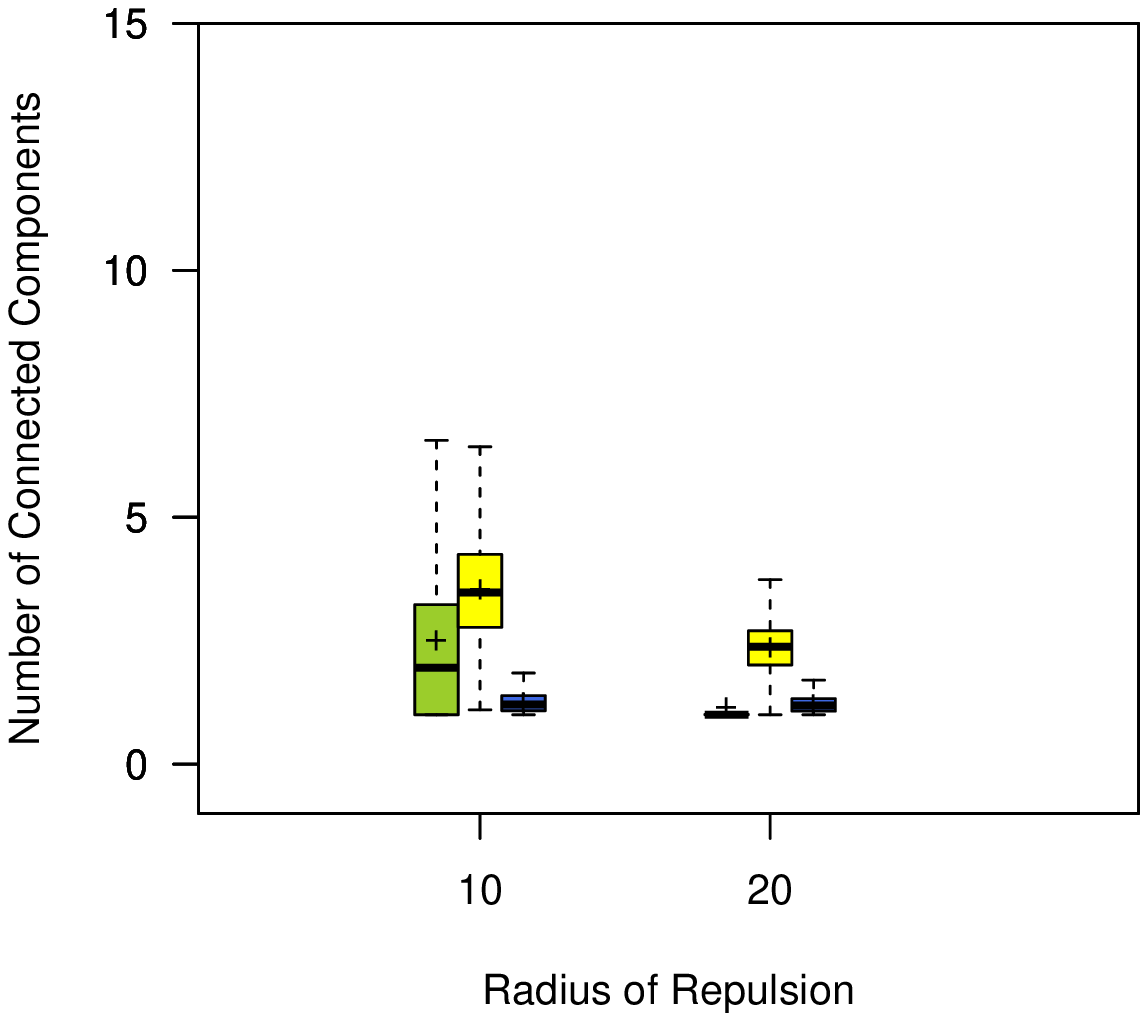}}
\hfill
\subfloat[{\it Search for a Goal}]{\includegraphics[trim=54 3 15 30,clip,scale=0.50, angle=270]{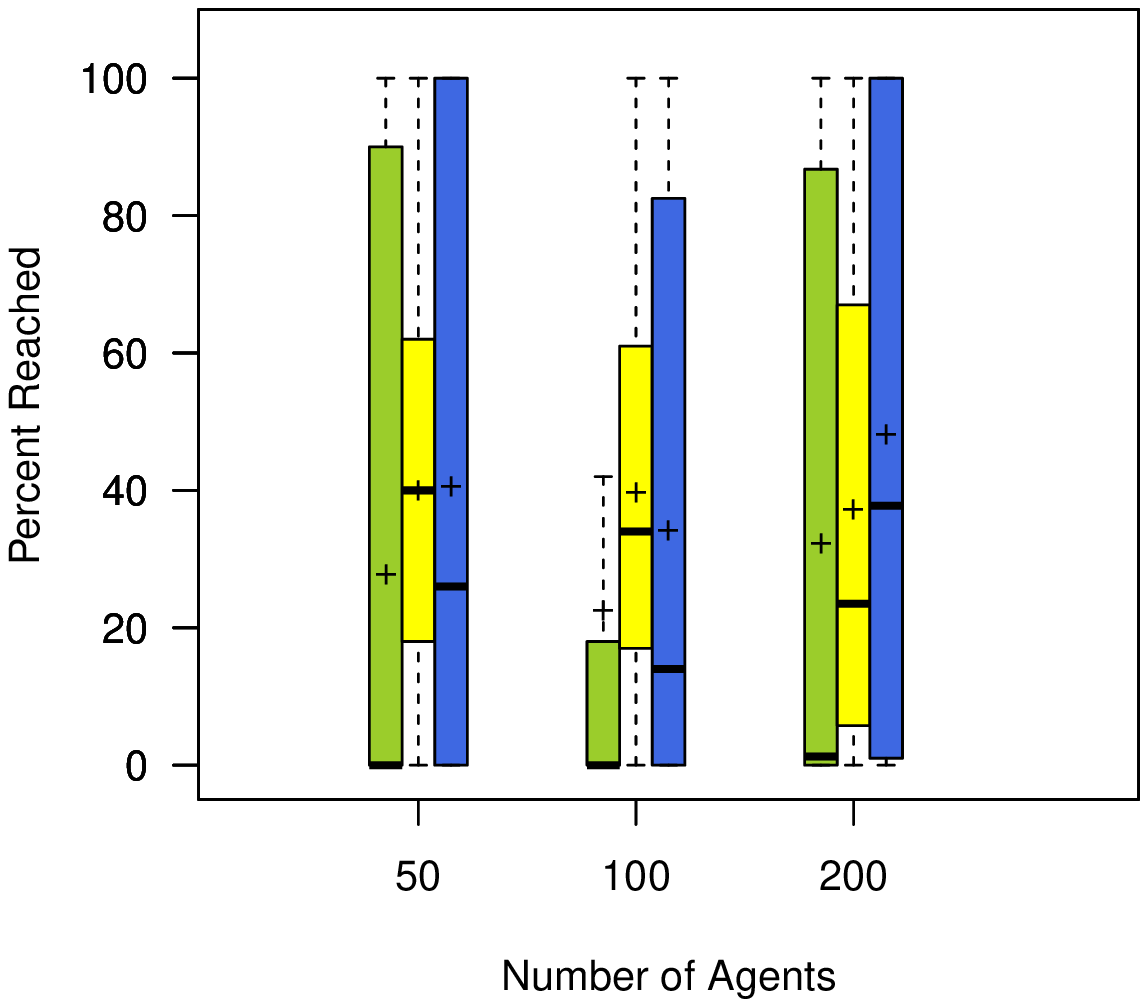}}
\hfill
\subfloat[{\it Search for a Goal}]{\includegraphics[trim=54 3 15 30,clip,scale=0.50, angle=270]{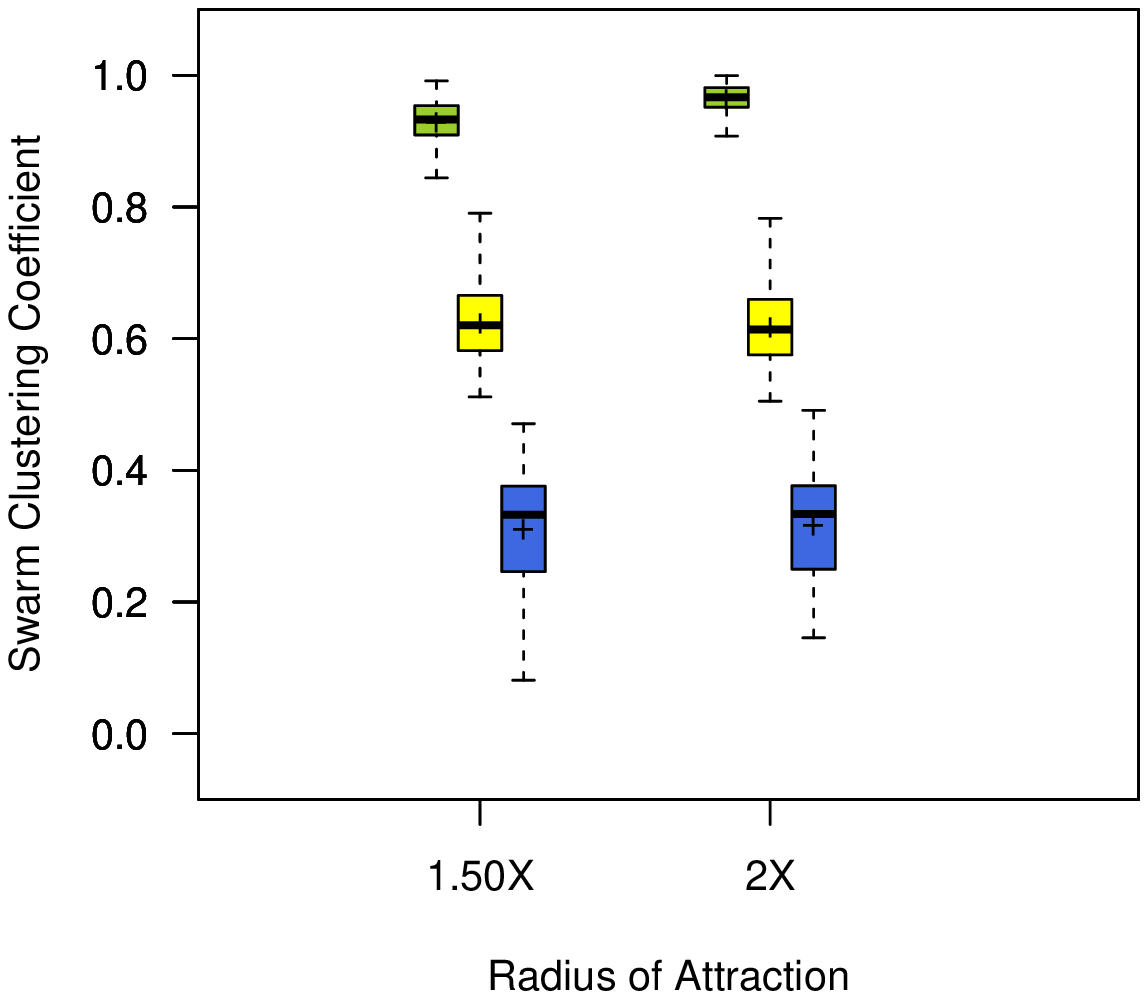}}
\hfill
\subfloat[{\it Rally}]{\includegraphics[trim=54 3 15 30,clip,scale=0.50, angle=270]{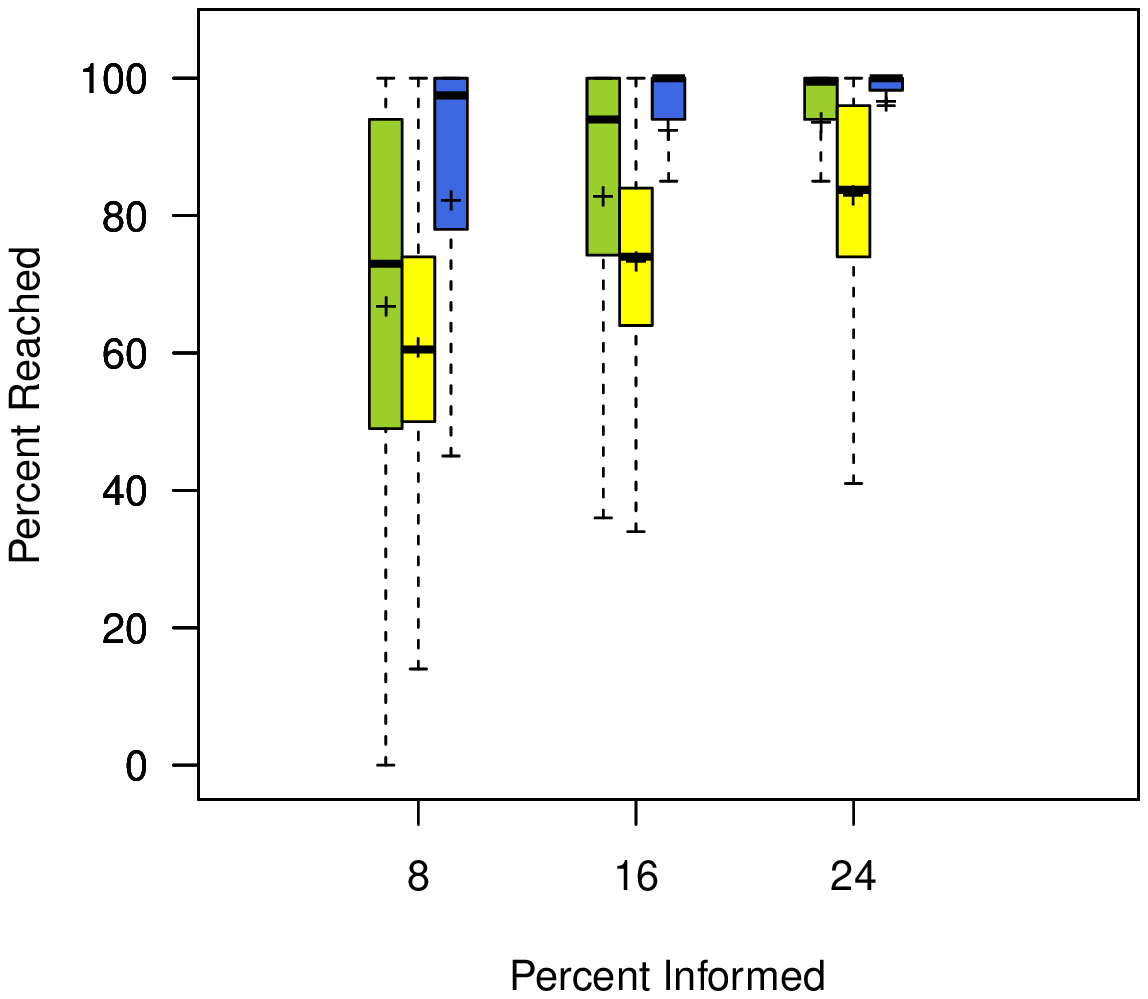}}
\hfill
\subfloat[{\it Rally}]{\includegraphics[trim=54 3 15 30,clip,scale=0.50, angle=270]{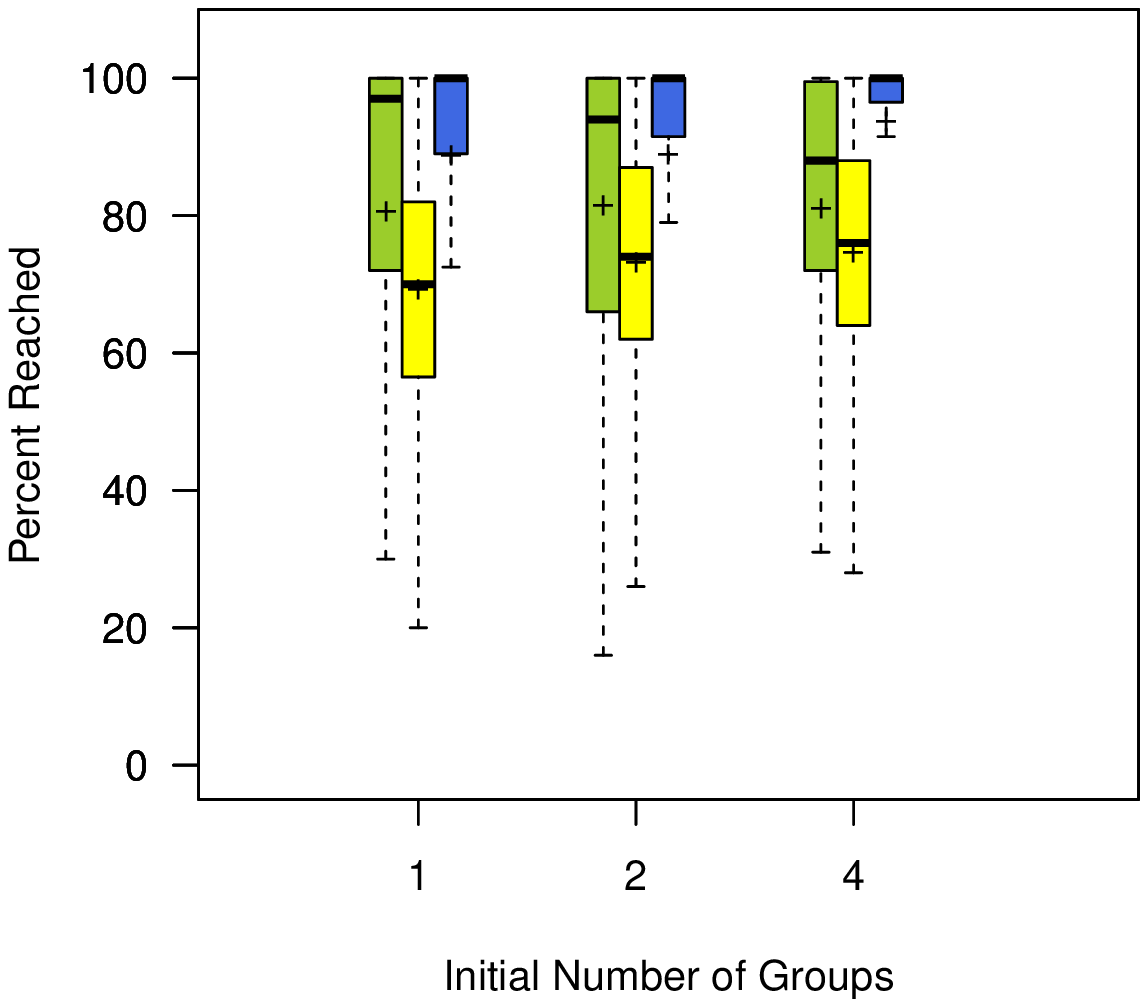}}
\hfill
\subfloat[{\it Disperse}]{\includegraphics[trim=54 3 15 30,clip,scale=0.50, angle=270]{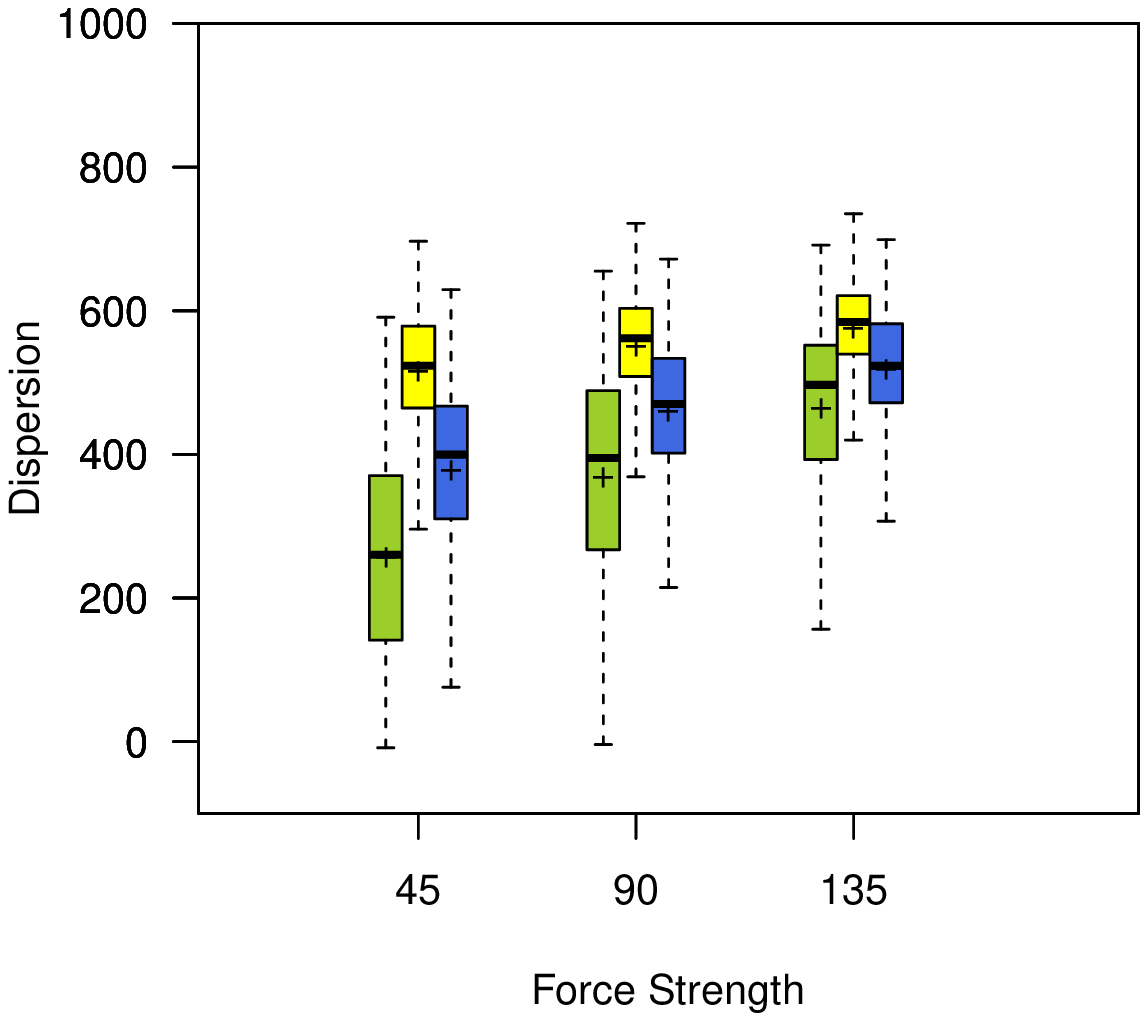}}
\hfill
\subfloat[{\it Disperse}]{\includegraphics[trim=54 3 15 30,clip,scale=0.50, angle=270]{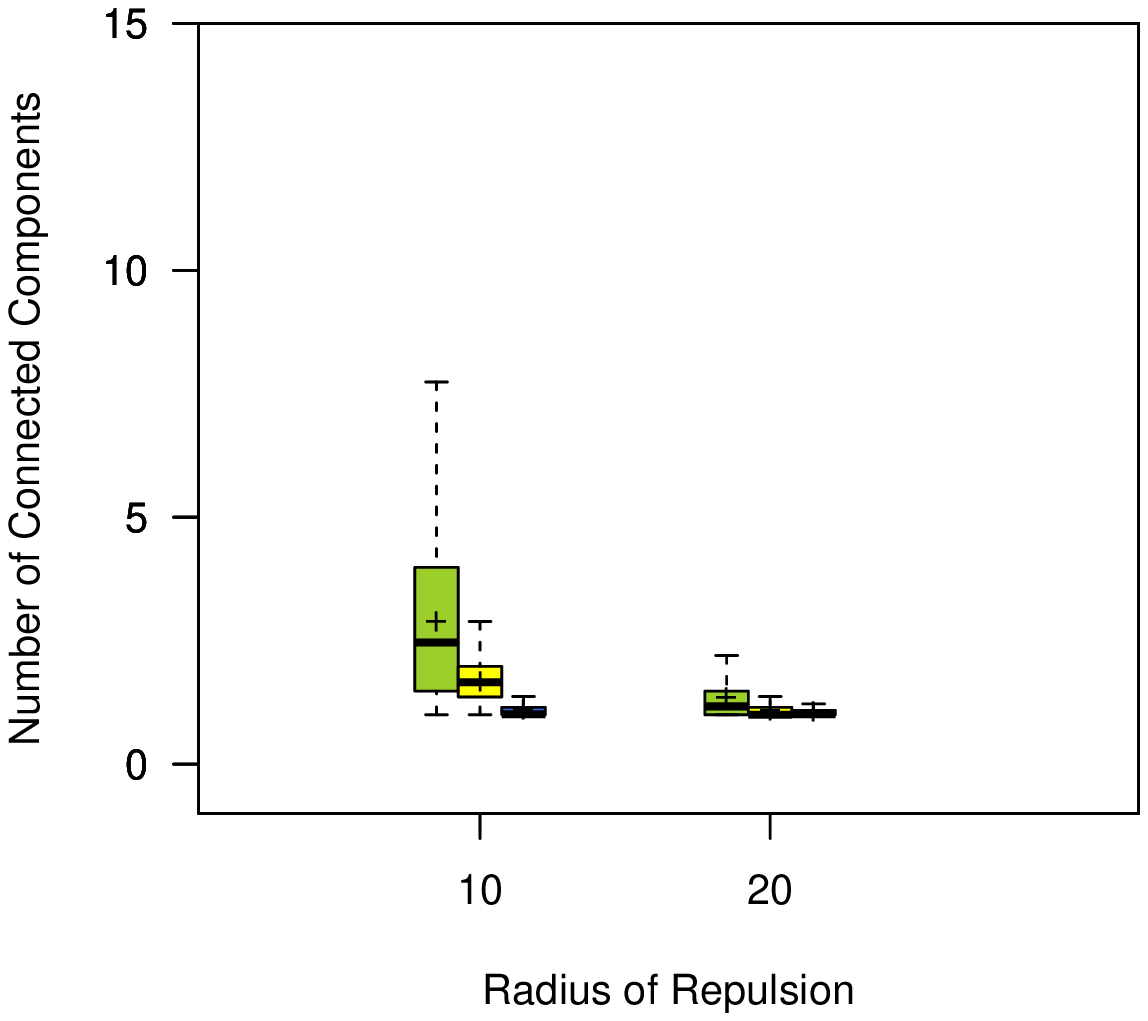}}
\hfill
\subfloat[{\it Avoid an Adversary}]{\includegraphics[trim=54 3 15 30,clip,scale=0.50, angle=270]{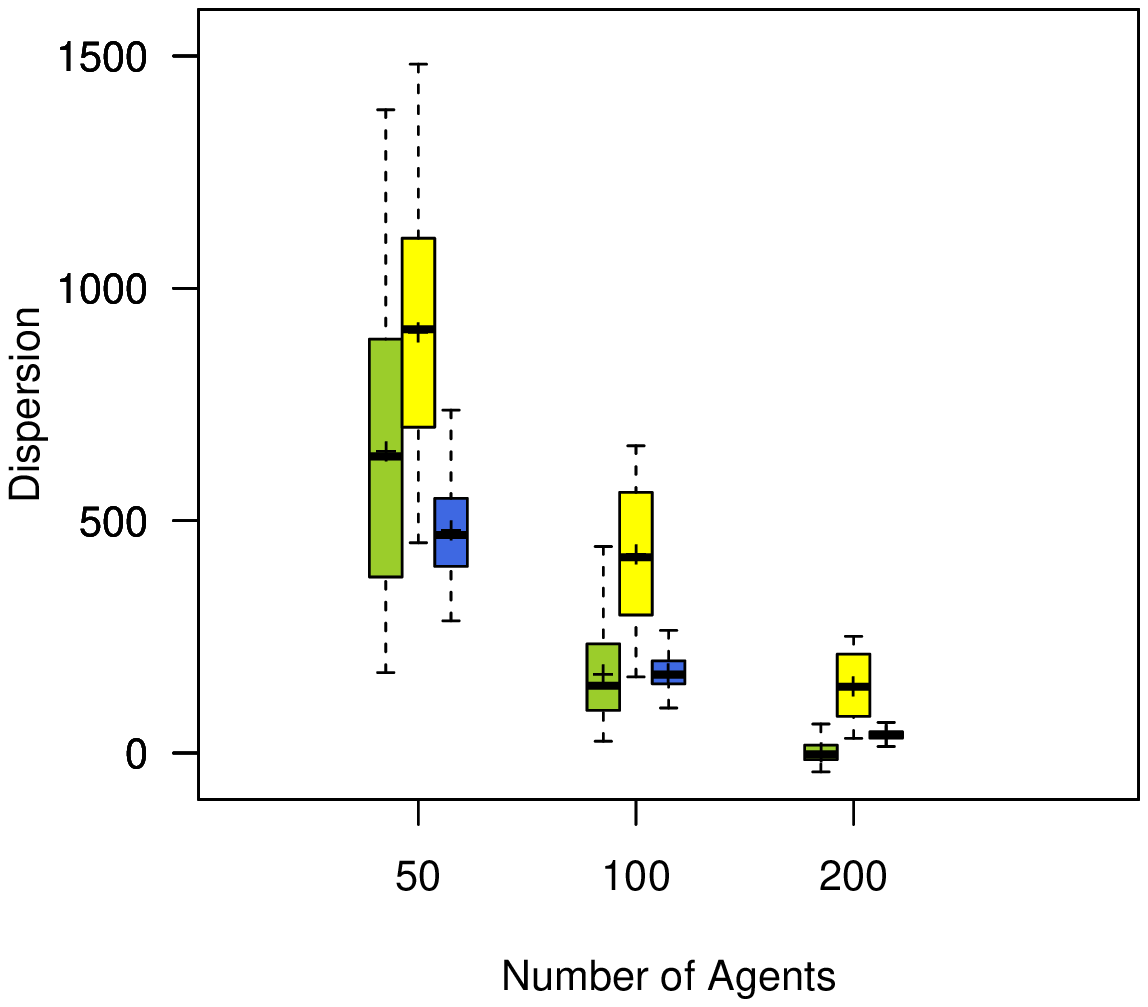}}
\hfill
\subfloat[{\it Avoid an Adversary}]{\includegraphics[trim=54 3 15 30,clip,scale=0.50, angle=270]{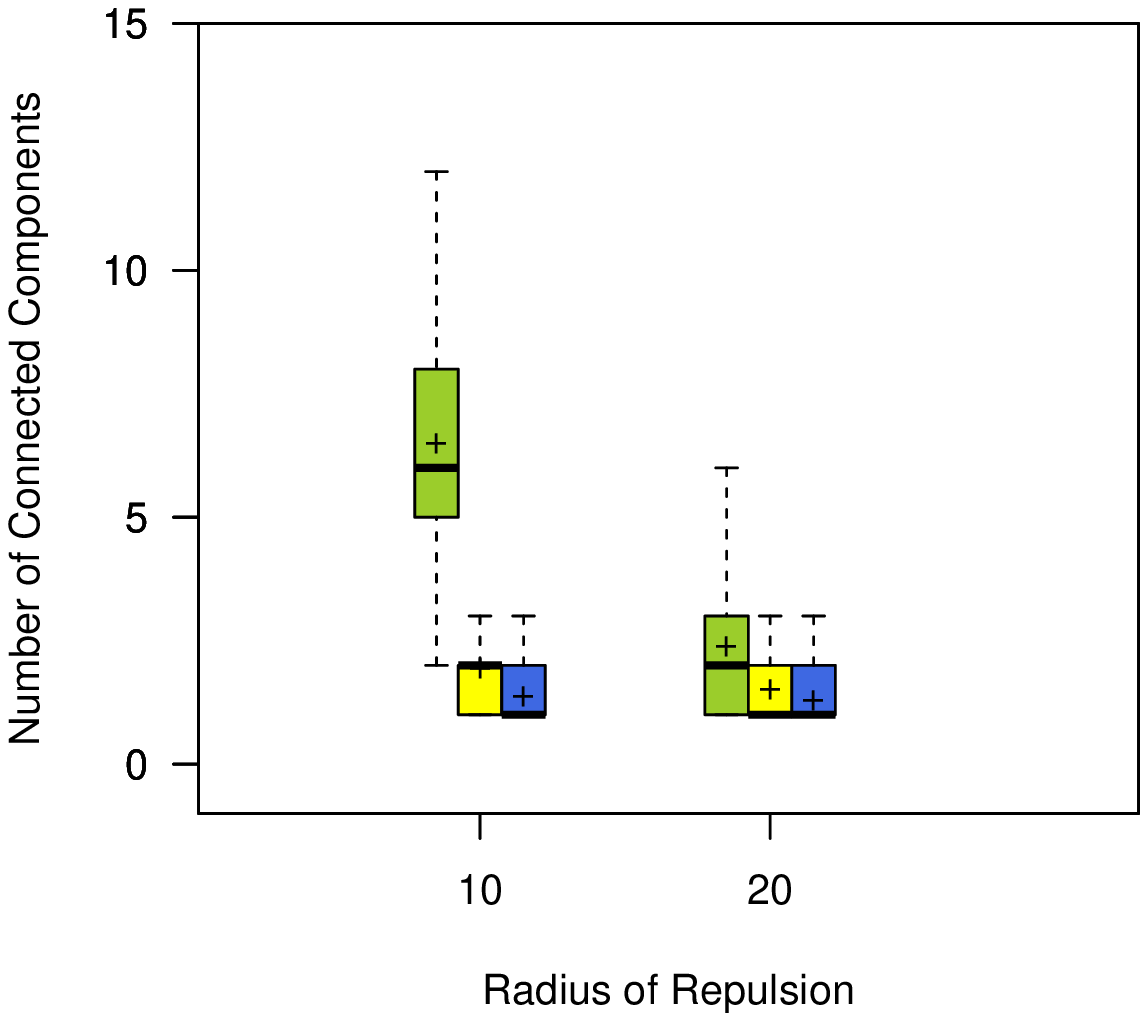}}
\hfill
\subfloat[{\it Follow}]{\includegraphics[trim=54 3 15 30,clip,scale=0.50, angle=270]{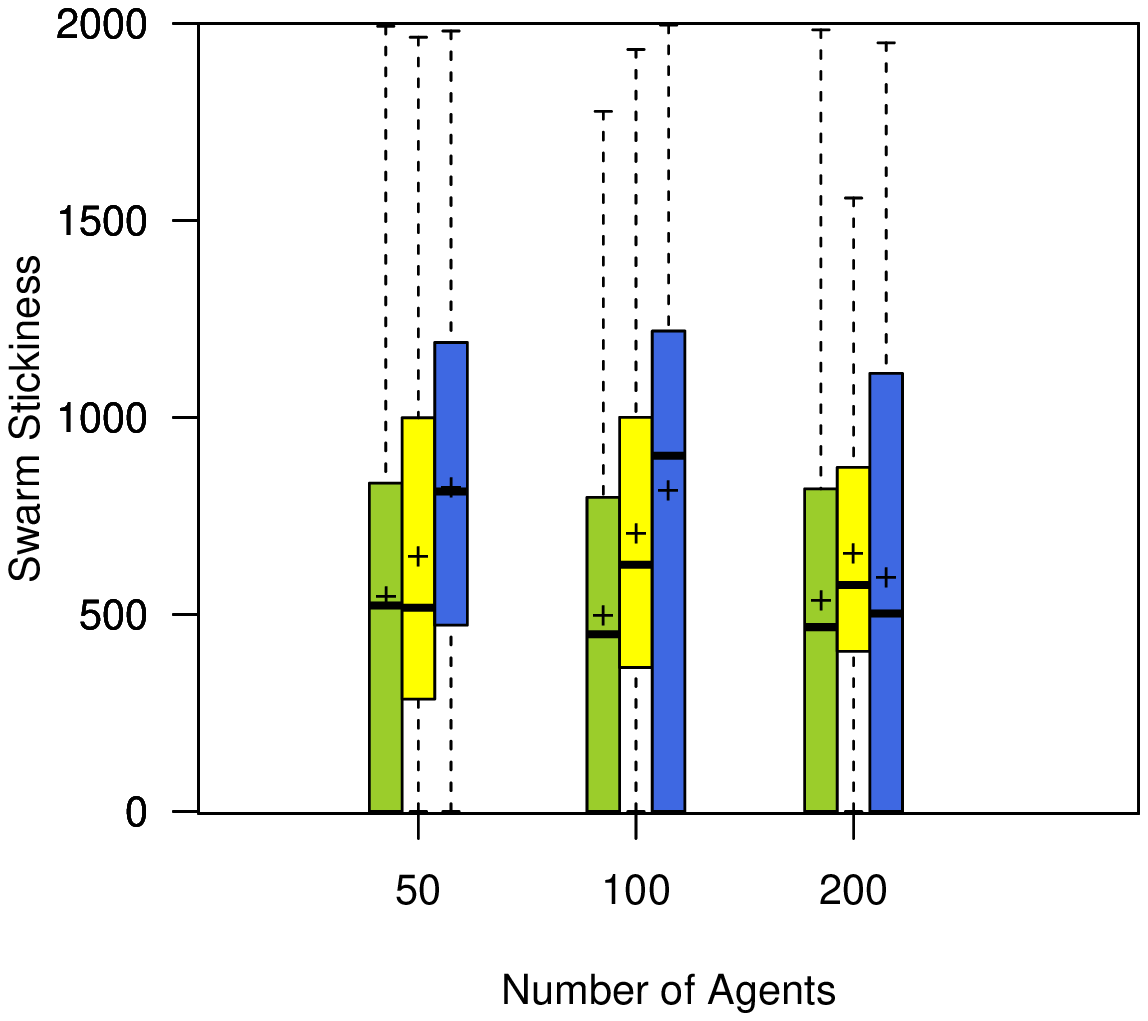}}
\hfill
\subfloat[{\it Follow}]{\includegraphics[trim=54 3 15 30,clip,scale=0.50, angle=270]{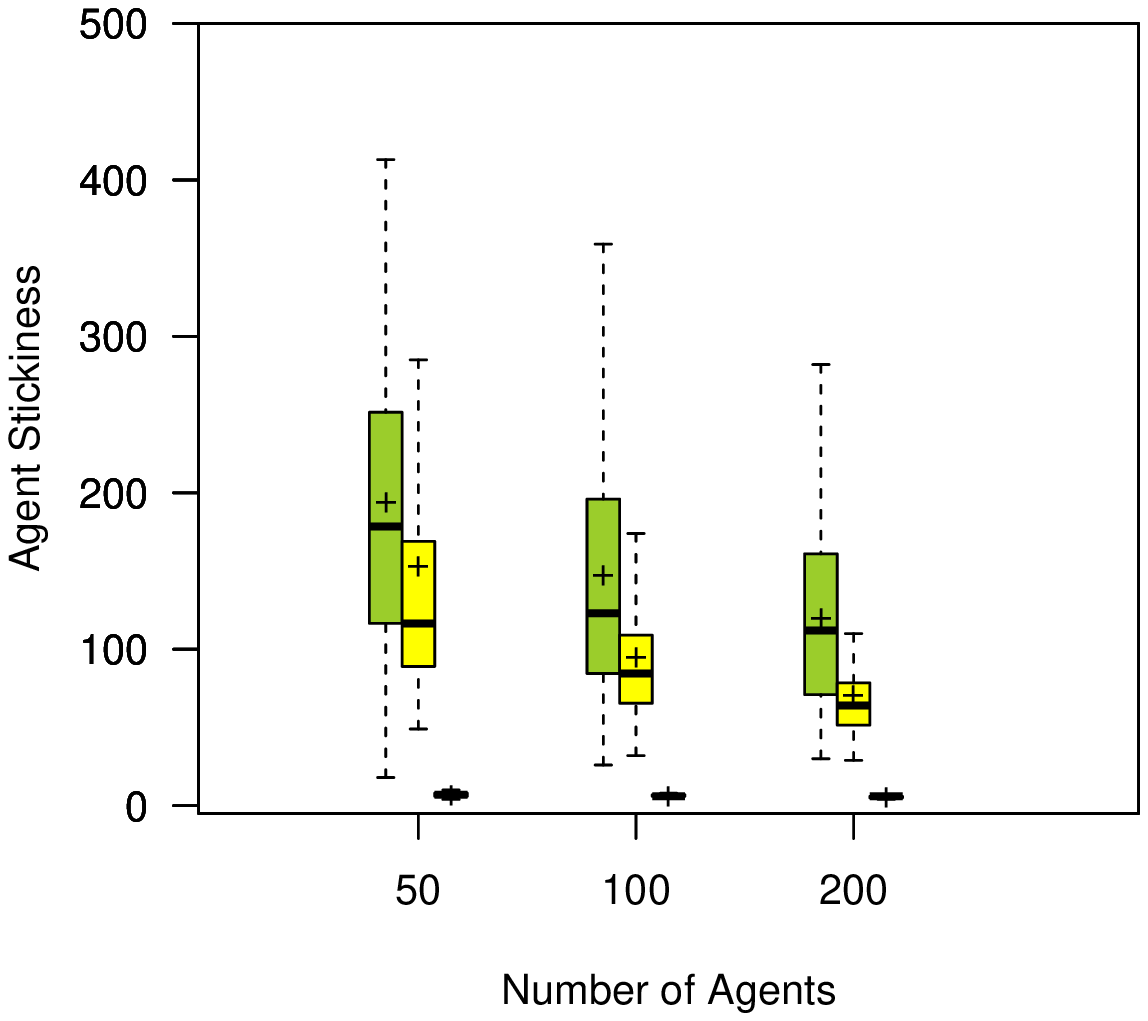}}
\caption{The performances of the three communication models further visualized by the experiment's additional factors.}
\label{fig:box_all}
\end{figure*}

The topological model had the highest mean {\bf percent found} ($M$ = 72.03, $SD$ = 19.61). An analysis of variance (ANOVA) showed that the effect of communication model on $PF$ was significant ($F_{8,5392}$ = 12,493.10, $p$ $<$ 0.001). Fisher's LSD post-hoc test revealed that the three models had significantly different performances compared to each other. The communication model had significant interactions with $r_{r}$ ($F_{2,5398}$ = 631.23, $p$ $<$ 0.001), $r_{o}$ ($F_{2,5398}$ = 160.75, $p$ $<$ 0.001), and $N_{o}$ ($F_{2,5398}$ = 228.48, $p$ $<$ 0.001). The model by $N_t$ interaction was not found to be significant (Figure \ref{fig:box_all}(a)). The visual model had a higher $PF$ ($M$ = 38.35, $SD$ = 17.98) than the metric model ($M$ = 31.66, $SD$ = 18.60) for most cases, except at the lowest values of the radii.

The visual model produced the lowest  {\bf number of connected components} ($M$ = 1.25, $SD$ = 0.22), whereas the topological model had the highest ($M$ = 2.97, $SD$ = 1.01). The metric model's mean $NCC$ was 1.85 ($SD$ = 1.45). ANOVA found that the effect of model type was significant ($F_{8,5392}$ = 7383.19, $p$ $<$ 0.001), and the post-hoc analysis of the pairwise differences showed that the models were significantly different from each other. The metric model at $r_{r} = 20$ yielded the lowest $NCC$ (Figure \ref{fig:box_all}(b)); otherwise, the visual model had the lowest $NCC$ across all $N$, $r_{o}$, and $r_{a}$ levels.

%


\subsection{Search for a Goal}
\label{sec:goto}


The overall mean {\bf percent reached} for this task was 35.95 ($SD$ = 39.38). The topological and visual models produced $PR$ means that were virtually identical (topological: $M$ = 39.08, $SD$ = 31.75; visual: $M$ = 41.10, $SD$ = 42.56). The metric model's mean $PR$ was 27.68 ($SD$ = 41.60). ANOVA found that the model type had a significant impact on $PR$ ($F_{6,1794}$ = 76.66, $p$ $<$ 0.001). There was no significant difference between the visual and topological models, but the metric model had a significantly lower $PR$ compared to the other models. There was no significant difference between the visual and topological models for $N=50$. The mean $PR$ was the highest at $N=100$ using the topological model; however, the visual model had the highest mean $PR$ at $N=200$ (Figure \ref{fig:box_all}(c)). No significant effect on $PR$ was identified due to the interactions between model and either $r_{r}$ or $r_{o}$.

A significant difference in {\bf latency} was found by an ANOVA between the models ($F_{6,1794}$ = 440.77, $p$ $<$ 0.001). $L$ for all three models were significantly different from each other in the post-hoc analysis (metric: $M$ = 637.79, $SD$ = 471.73; topological: $M$ = 864.99, $SD$ = 290.20; visual: $M$ = 438.73, $SD$ = 487.99). The metric and topological models typically had a median of 1000 for most cases; visual median was 31.00.

The mean {\bf swarm clustering coefficient} was lowest in the visual model ($M$ = 0.31, $SD$ = 0.07) and highest in the metric model ($M$ = 0.95, $SD$ = 0.03). The topological model's mean was 0.62 ($SD$ = 0.06). The effect of model type on $SCC$ was significant ($F_{6,1794}$ = 1810, $p$ $<$ 0.001). Post-hoc analysis revealed significant pairwise differences between the models. The median $SCC$ for all communication models were generally close to the means across all parameters and associated values. The interquartile ranges were typically tight (Figure \ref{fig:box_all}(d)), with only a few cases where the maximum value of one model overlapped with the minimum value of another.


\subsection{Rally}
\label{sec:rally}


The overall mean {\bf percent reached} ($M$ = 81.40, $SD$ = 23.08) was higher compared to the prior task. ANOVA showed that the model type had a significant effect on $PR$ ($F_{8,5392}$ = 1175.31, $p$ $<$ 0.001). Post-hoc analysis revealed that all three models' performance was significantly different from one another. The mean $PR$ was highest for the visual model ($M$ = 90.55, $SD$ = 21.17), and the metric model ($M$ = 81.17, $SD$ = 25.06) outperformed the topological model ($M$ = 72.48, $SD$ = 18.92). Model type significantly interacted with $p_i$ ($F_{4,5396}$ = 90.93, $p$ $<$ 0.001); however, the interaction with $g$ was not found to be significant (Figures \ref{fig:box_all}(e)-(f)).

ANOVA showed that there were significant differences across the models in the mean {\bf direct influence} exhibited by informed agents on the remainder of the swarm ($F_{8,5392}$ = 14997.31, $p$ $<$ 0.001). The metric model produced the highest $DINF$ ($M$ = 0.90, $SD$ = 0.12), whereas, the topological model had the lowest (topological: $M$ = 0.70, $SD$ = 0.17; visual: $M$ = 0.81, $SD$ = 0.13). A post-hoc test using the Fisher's LSD test revealed significant pairwise differences between the models.


\subsection{Disperse and Avoid an Adversary}
\label{sec:disperseavoid}


The effect of the communication model on the {\bf dispersion} was significant during the {\it Disperse} ($F_{8,5392}$ = 9232.53, $p$ $<$ 0.001) and {\it Avoid an Adversary} tasks ($F_{4,596}$ = 492.82, $p$ $<$ 0.001). The topological model had the highest $D$ for both tasks (see Table \ref{table:disperseavoid}). ANOVA found the model by $s$ interaction to be significant for the {\it Disperse} task ($F_{8,5392}$ = 996.64, $p$ $<$ 0.001). The topological model at the lowest $s$ value and the visual model at the highest $s$ value had comparable performances (Figures \ref{fig:box_all}(i)). There were significant interactions between model and $N$ for the {\it Disperse} ($F_{8,5392}$ = 1112.47, $p$ $<$ 0.001) and {\it Avoid} ($F_{4,596}$ = 118.32, $p$ $<$ 0.001) tasks. Generally during the {\it Disperse} task, $D$ increased in $N$, but for all models, the opposite occurred in the {\it Avoid} task (Figures \ref{fig:box_all}(i)). Comparing the visual and metric models for the {\it Avoid} task revealed that at $N$ = 50, the visual model was significantly lower; at $N$ = 100, no significant difference was found; and at $N$ = 200, the visual model was significantly higher.

The metric model produced the greatest {\bf number of connected components}, followed by the topological model, and then the visual model. The communication model's effect on $NCC$ was significant for both the {\it Disperse} ($F_{8,5392}$ = 4224.10, $p$ $<$ 0.001) and {\it Avoid} ($F_{4,596}$ = 1638.76, $p$ $<$ 0.001) tasks. The metric and visual models deliver comparable $NCC$ at $r_{r}=20$ for both tasks (Figures \ref{fig:box_all}(h) and \ref{fig:box_all}(j)).

\begin{table}
\caption{The Disperse and Avoid an Adversary tasks' descriptive statistics (Mean (SD)).}
\label{table:disperseavoid}       
\begin{tabular}{llrrr}
\hline\noalign{\smallskip}
Task & Model & $D$ & $NCC$ & $I$\\
\noalign{\smallskip}\hline\noalign{\smallskip}
& Metric & 364.52 (160.87) & 2.14 (1.51) & 0.66 (1.08)\\
& &  & & \\
Disperse & Topological & 548.01 (75.15) & 1.39 (0.45) & 0.00 (0.00)\\
& &  & & \\
& Visual & 452.81 (121.02) & 1.12 (0.23) & 0.16 (0.39)\\
& & & & \\
\hline
& & & & \\
& Metric & 275.61 (334.85) & 4.46 (2.78) & 1.19 (1.38)\\
& & & & \\
Avoid & Topological & 493.92 (356.32) & 1.75 (0.79) & 0.00 (0.00)\\
& & & & \\
& Visual & 232.03 (196.64) & 1.35 (0.58) & 0.33 (0.54)\\
\noalign{\smallskip}\hline
\end{tabular}
\end{table}
%


\subsection{Follow}
\label{sec:follow}


The effect of model type on  {\bf swarm stickiness} was significant ($F_{4,596}$ = 26.89, $p$ $<$ 0.001), and the visual model's $SSTK$ was the highest compared to the other models (metric: $M$ = 528.75, $SD$ = 497.85; topological: $M$ = 671.92, $SD$ = 476.27; visual: $M$ = 746.34, $SD$ = 604.77). The metric model produced the lowest $SSTK$ for all $N$ (Figure \ref{fig:box_all}(k)), yet at $N$ = 200, it had a comparable performance to the visual model.

The metric model had the highest {\bf agent stickiness} (metric: $M$ = 154.23, $SD$ = 93.57; topological: $M$ = 106.48, $SD$ = 91.92; visual: $M$ = 6.45, $SD$ = 1.01). ANOVA revealed that model type had a significant impact on $ASTK$ ($F_{4,596}$ = 925.26, $p$ $<$ 0.001), and a post-hoc analysis revealed significant pairwise differences between the models. $ASTK$ decreased in $N$ for all models (Figure \ref{fig:box_all}(l)).

ANOVA revealed the main effect of model type on the leader's {\bf influence} on the swarm to be significant ($F_{4,596}$ = 6068.82, $p$ $<$ 0.001). $INF$ was highest for swarms using the visual model ($M$ = 0.94, $SD$ = 0.06). The metric and topological swarms yielded much lower $INF$ compared to the visual model (metric: $M$ = 0.26, $SD$ = 0.23; topological: $M$ = 0.39, $SD$ = 0.21). The pairwise differences between the three models was significant, according to the post-hoc analysis. The visual model's high $INF$ was further observed across all values of the additional factors.


\section{Discussion and Conclusions}
\label{sec:discussion}


The presented research focuses on a general hypothesis that the selection of a communication model impacts a swarm's task performance. Six swarm robotics tasks were investigated for the three most predominant communication models found in the biological swarm literature. The primary finding is that different tasks benefit from different models, and as such, the task by communication model pairing is an important dimension in the effective design of artificial swarms.

No single model outperformed the others across all the tasks; however, some general trends emerged within the limited task design considerations. The visual model was more beneficial in tasks that required the swarm to move to a particular area (see Table \ref{table:recommendation}). The two tasks that had this transport-like flavor were the {\it Search for a Goal} and {\it Rally} tasks. The topological model was better at enduring a force directed toward the swarm, as is the case with the {\it Disperse} and {\it Avoid an Adversary} tasks.

The visual model, with its {\it potentially} long communication links was better able to keep the swarm together, which resulted in the lowest number of connected components. This tendency also led to the model fairing poorly when exploring the environment during the {\it Search for Multiple Targets} task.

Agents favorably oriented and not occluded had a higher chance of establishing long-range links using the visual model. Thus, agents were more likely to receive the goal's location ({\it Search for a Goal}) or be influenced ({\it Rally}) by an informed agent. Any occurrence of a long-range link in the network, regardless of how infrequent, acted as a ``short-cut" \cite{linkedBook} for transferring information. Despite this advantage, at the lowest and highest attraction values, the metric model had comparable and lower latency, respectively, compared to the visual model.

Communication links in the metric and topological models were unaffected by occlusions, a factor that yielded sparser networks for the visual model \cite{strandburgVisual}. A low clustering can be disadvantageous in noisy environments without the benefit of redundant links. Given a noisy environment with a low requirement on the percentage of agents reaching goal, the metric model is recommended.

\begin{table}[t]
\caption{The recommended communication models by task.}
\label{table:recommendation}       
\begin{tabular}{lc}
\hline\noalign{\smallskip}
Task & Recommended Model\\
\noalign{\smallskip}\hline\noalign{\smallskip}
Search for Multiple Targets & Topological\\
Search for a Goal & Visual\\
Rally & Visual\\
Disperse & Topological\\
Avoid an Adversary & Topological\\
Follow & Visual\\
\noalign{\smallskip}\hline
\end{tabular}
\end{table}

A high dispersion in some species may serve to confuse a predator from singling out a particular swarm agent \cite{balleriniTopological}. Thus, if a higher dispersion is preferred, the topological model was the best model for the {\it Avoid an Adversary} task. The model produced the highest dispersion, low connected components, and no isolated components. A limitation of the {\it Avoid} task's design was the use of a singe adversary approaching in a pre-defined motion, rather than a (coordinated) attack from multiple adversaries.

Swarm agents in frontal positions influenced agents behind them to follow the {\it Follow} task's leader, in a cascading effect, when using the visual model. The leader was lost multiple times during a trial, a drawback of the model. The metric model is a better choice for {\it persistent} tracking (and if tracking by a small fraction of the swarm is tolerable). The metric model tended to break the swarm into numerous, stable clusters, one of which typically contained the leader.

The implemented model parameter values agreed with reported behavior, and provides connections to the biological swarm literature. Couzin {\it et al.} \cite{couzinRadii} showed that $r_{r}$ does not have an effect on the transitions between different swarm movement patterns. Rather, the relative sizes of $r_{o}$ to $r_{r}$, and $r_{a}$ to $r_{o}$ produce the transitions. Simulated swarms, for instance, rotate in a torus when the $r_{o}/r_{r}$ ratio is relatively low and the $r_{a}/r_{o}$ ratio is relatively high. Presented results for the {\it Search for a Goal} task conform to the $r_{r}$ finding. The duration of this task resulted in trials that demonstrated swarm movement patterns described by Couzin {\it et al.} \cite{couzinRadii}, and the performance was not impacted by the choice in $r_{r}$. A mapping of movement types to performance was beyond the scope of this work.

The scope of the reported research does not follow the so-called prescriptive agenda \cite{shammaQuestion,shohamQuestion}, where the values of the model parameters are free design choices; thus, $d_{vis}$ and $\phi$ are not varied, for instance. This line of inquiry will become necessary when specific platforms attempt to adopt the models (e.g., s-bots \cite{dorigoSwarmbots} are equipped with vision sensors). The metric model, for instance, can be realized with omni-directional antennas, as well as infrared LED sensors; however, the LED range in Kilobots is only $10$cm \cite{nagpalKilobots}. Similarly, for the topological model, which can be implemented using band-limited communication channels \cite{goodrichTR}, the infrared-based, band-limited platforms such as the r-one \cite{jamesRone}, achievable $n_{top}$ will be constrained by the maximum communication range.

Aside from task specific limitations, one of the general limitations of the overall evaluation is the focus on individual tasks. For instance, each experiment is associated with performing a unique task. Additional analysis over task combinations is required to fully support the general hypothesis. However, the presented results provide preliminary evidence that support the general hypothesis.

%
%
%
%
%

\section*{Acknowledgment}

This material is based upon research supported (in part) by the U.S. Office of Naval Research award \#N000141210987.

%

\ifCLASSOPTIONcaptionsoff
  \newpage
\fi






\end{document}